\documentclass[11pt,a4paper]{article}

\usepackage[T1]{fontenc}
\usepackage[utf8]{inputenc}
\usepackage[margin=2.2cm]{geometry}
\usepackage{microtype}
\usepackage{parskip}
\usepackage{amsmath,amssymb,amsthm,mathtools}
\usepackage{bm}
\usepackage{graphicx}
\graphicspath{{figures/}}
\usepackage{booktabs}
\usepackage{array}
\usepackage{longtable}
\usepackage{caption}
\usepackage{xcolor}
\usepackage{mdframed}
\usepackage{hyperref}
\hypersetup{
  colorlinks=true,
  linkcolor=blue!50!black,
  citecolor=blue!50!black,
  urlcolor=blue!60!black,
  pdftitle={An Isotropy-Preserving Spectral Cap for Muon},
  pdfauthor={Jiachun Li},
}

\newcommand{\code}[1]{\texttt{\small #1}}
\newcommand{\Fro}[1]{\|#1\|_F}
\newcommand{\spec}[1]{\|#1\|_2}
\newcommand{\msign}{\operatorname{msign}}
\newcommand{\tr}{\operatorname{tr}}

\theoremstyle{definition}
\newtheorem{assumption}{Assumption}
\theoremstyle{remark}

\newmdenv[
  linecolor=orange!55!black,
  backgroundcolor=orange!6,
  linewidth=0.8pt,
  roundcorner=3pt,
  skipabove=6pt,
  skipbelow=6pt,
  innertopmargin=6pt,
  innerbottommargin=6pt
]{caveatbox}

\title{\bfseries An Isotropy-Preserving Spectral Cap for Muon:\\
Theory and Three Case Studies\\[4pt]
\large\normalfont(Preliminary report --- comments welcome)}
\author{Jiachun Li}
\date{2026-06 (draft)}

\begin{document}
\maketitle

\begin{caveatbox}
\textbf{Status of this report.} This is a \emph{preliminary} technical report. The theory in
Section~\ref{sec:theory} rests on a fairly strong idealisation --- \emph{exact scale invariance}
of the loss with respect to weight rescaling (Assumption~\ref{ass:scale}). Real
normalisation-heavy networks are only \emph{approximately} scale invariant, so several of the
theoretical claims should be read as clean limiting statements rather than established facts.
We give a first empirical sanity check in Section~\ref{sec:finding1:cpcg}, but confirming (or
refuting) the assumption and its consequences will require substantially more experiments. We
share the report at this stage precisely to invite feedback; corrections and counter-examples
are very welcome.
\end{caveatbox}

\begin{abstract}
This report has four parts. \textbf{Part~1 (theory)} proposes a single explanatory framework.
In a scale-invariant network the loss is insensitive to the scale $\Fro{W}$ of a weight matrix.
Under this idealisation, plain SGD carries a built-in $1/\Fro{W}$ ``brake'' on its update size,
whereas the matrix-sign step used by Muon removes that brake, so both the Frobenius norm and the
spectral norm drift outward faster. We then note that the perturbation of the spectral norm has a
\emph{non-negative second-order term} (a convexity effect): even after the first-order
top-direction component of an update is projected away, the weight can still learn through three
mechanisms --- growth of non-top singular directions, rotation of the top direction, and
top-direction switching. This is the core reason a \emph{spectral cap} can control the spectrum
without freezing training.

\textbf{Parts~2, 3 and 4} test the idea on three concrete systems: a nanoGPT feed-forward
projection, a 64-expert MoE router, and the query/key projections of a FlashAttention block run
in bf16. For each system we state precisely what the weight $W$, the input activation $X$, and the
covariance $K_X$ are, together with the algorithm and cost of finding the top singular pair
$(u_1, v_1)$.

\medskip
\noindent\textbf{The three case studies at a glance:}
\begin{center}
\small
\begin{tabular}{p{0.05\textwidth}p{0.15\textwidth}p{0.19\textwidth}p{0.21\textwidth}p{0.28\textwidth}}
\toprule
Part & System & $W$ (matrix the cap controls) & $X$ (input activation) & Main finding \\
\midrule
2 & nanoGPT FFN & \code{mlp.c\_proj} $\in\mathbb{R}^{384\times 1536}$ & GELU hidden, $X\in\mathbb{R}^{M\times 1536}$ & data-cap lowers the top-share of the c\_proj output covariance $K_Y$ from $0.74$ to $0.30$ with no change in val loss \\
3 & 64-expert MoE router & router $\in\mathbb{R}^{64\times 384}$ & post-RMSNorm token, $X\in\mathbb{R}^{M\times 384}$ & selection cap pushes the top-share of $K_{z+b}$ to $0.12$--$0.52$ across all six layers; without it, layer~0 collapses to a single dominant expert (top-share $0.995$) \\
4 & FlashAttention Q/K & $W_Q^{(h)}, W_K^{(h)}\in\mathbb{R}^{64\times 768}$ (per head) & token embedding, $X\in\mathbb{R}^{M\times 768}$ & at iteration 86K a single head (L1~H2) reaches $\lambda_{\max}=133$K and is caught in real time by the cap; the uncapped baseline crashes in the same window \\
\bottomrule
\end{tabular}
\end{center}
\end{abstract}

\tableofcontents
\clearpage

\section{Theory: a spectral cap and isotropy under Muon}
\label{sec:theory}

\subsection{Setup and goal}
\label{sec:theory:goal}

Consider a matrix weight $W \in \mathbb{R}^{m\times n}$ that maps an input activation
$X \in \mathbb{R}^{N\times n}$ linearly to the output $Y = X W^\top \in \mathbb{R}^{N\times m}$.
We care about the \emph{second moments at the input and output}:
\[
  K_X := \tfrac{1}{N}X^\top X \in \mathbb{R}^{n\times n},\qquad
  K_Y := \tfrac{1}{N}Y^\top Y = W K_X W^\top \in \mathbb{R}^{m\times m}.
\]
$K_X$ describes how the input energy is distributed over directions in the $n$-dimensional feature
space, and $K_Y$ is the output energy actually seen by the downstream module (attention, softmax,
or another FFN block).

For the derivations in this section we take the simplifying case
\[
  K_X \approx I,
\]
and defer the general $K_X \ne I$ case to Section~\ref{sec:theory:remark}. Under $K_X = I$,
\[
  K_Y = W W^\top,\qquad
  \lambda_1(K_Y) = \sigma_1(W)^2,\qquad
  \tr K_Y = \Fro{W}^2,
\]
so the natural ``top-share'' quantity is
\[
  q := \frac{\lambda_1(K_Y)}{\tr K_Y} = \frac{\sigma_1(W)^2}{\Fro{W}^2}
     = \frac{1}{\text{stable rank}(W)}.
\]
The failure mode we want to avoid is $K_Y$ degenerating towards (approximate) rank one: a single
top eigenvalue $\lambda_1(K_Y)$ much larger than the rest, i.e.\ $q \to 1$. This shows up as a
concrete physical failure in each of the three case studies:
\begin{itemize}\itemsep0pt
\item \textbf{nanoGPT FFN}: $K_Y$ is the covariance the FFN writes into the residual stream; a high
top-share means the FFN writes almost entirely along one direction.
\item \textbf{MoE router}: $K_Y$ is the covariance of the selection score over tokens; top-share
$\to 1$ means all tokens route to the same expert (rank-1 collapse).
\item \textbf{FlashAttention Q/K}: $\lambda_{\max}(K_Q^{(h)})$ directly controls the magnitude of the
attention scores; when it is too large, the bf16 mantissa can no longer separate multiple near-maximal
entries in a row, softmax rounds in a biased way, and training destabilises.
\end{itemize}
So we use $\lambda_1(K_Y)$ as the failure indicator for isotropy and try to \emph{stop it from
growing too fast under Muon updates}. That is what the spectral cap does. The two subsections that
follow establish the two structural claims:
\begin{enumerate}\itemsep0pt
\item Muon and SGD affect $\Fro{W}$ and $\spec{W}$ differently
(Sections~\ref{sec:theory:scale}--\ref{sec:finding1});
\item a spectral cap on $\spec{W}$ can control anisotropy at a low order without freezing the weight
(Sections~\ref{sec:finding2}--\ref{sec:theory:entropy}).
\end{enumerate}

\subsection{The scale-invariance assumption}
\label{sec:theory:scale}

Many normalisation-heavy networks (pre-LN / RMSNorm transformers, and other dense layers whose output
is normalised) are almost insensitive to the overall scale of a weight matrix: RMSNorm/LN divides out
$\Fro{W}$, so $L(cW) \approx L(W)$ for $c > 0$. We idealise this into an exact statement.

\begin{assumption}[Exact scale invariance]\label{ass:scale}
For the weight matrices we study, the loss is invariant under positive rescaling:
\[
  L(cW) = L(W)\qquad\text{for all } c > 0.
\]
\end{assumption}

\begin{caveatbox}
Assumption~\ref{ass:scale} is a \textbf{strong idealisation} and is the main load-bearing assumption
of the theory. In practice normalisation makes networks only \emph{approximately} scale invariant
(biases, weight decay, residual connections, and the last unnormalised layer all break it). Every
consequence below --- the $1/\Fro{W}$ brake, the $t^{1/4}$ vs $t^{1/2}$ norm-growth rates, and the
orthogonality $\langle W, G\rangle_F \approx 0$ --- inherits this caveat. We give a first empirical
check of $\langle W, G\rangle_F \approx 0$ in Section~\ref{sec:finding1:cpcg}, but whether the
assumption holds tightly enough for the quantitative predictions to survive is an open question that
needs more measurement across architectures and training stages.
\end{caveatbox}

\subsubsection{Corollary~B (weak form): the gradient is orthogonal to $W$}

Differentiating $L(cW) = L(W)$ in $c$:
\[
  \frac{d}{dc}L(cW) = \langle W, \nabla_W L(cW)\rangle_F = 0\qquad\text{for all } c > 0.
\]
Evaluating at $c = 1$ gives
\[
  \boxed{\;\langle W, \nabla_W L(W)\rangle_F = 0.\;}
\]
Geometrically: $W$ (the radial direction, which is exactly the scale-invariant direction) receives no
raw-gradient component. The raw gradient $\nabla_W L$ lives in the subspace orthogonal to $\Fro{W}$.

\subsubsection{Corollary~A (scale form): the gradient scales like $1/\Fro{W}$}

Scale invariance means $L$ depends only on the direction $\Theta := W/\Fro{W}$, so we can write
$L(W) = \ell(\Theta)$. With $r := \Fro{W}$, the differential of $r$ along a perturbation $H$ is
\[
  dr[H] = \Big\langle \tfrac{W}{\Fro{W}}, H\Big\rangle_F = \langle \Theta, H\rangle_F,
\]
using $r = \sqrt{\langle W, W\rangle}$. Hence, for $\Theta = W/r$,
\[
  d\Theta[H] = \frac{H}{r} - \frac{W}{r^2}dr[H]
             = \frac{1}{r}\big(H - \Theta\langle \Theta, H\rangle_F\big)
             = \frac{1}{r}\,P_{\Theta^\perp}H,
\]
where $P_{\Theta^\perp}H := H - \Theta\langle \Theta, H\rangle_F$ is the projection onto the
Frobenius-orthogonal complement of $\Theta$. By the chain rule on $L(W) = \ell(\Theta(W))$,
\[
  \langle \nabla_W L, H\rangle_F
  = \langle \nabla_\Theta\ell, d\Theta[H]\rangle_F
  = \tfrac{1}{r}\langle \nabla_\Theta\ell, P_{\Theta^\perp}H\rangle_F
  = \tfrac{1}{r}\langle P_{\Theta^\perp}\nabla_\Theta\ell, H\rangle_F
\]
(using that $P_{\Theta^\perp}$ is a self-adjoint projection). Therefore
\[
  \boxed{\;\nabla_W L(W) = \frac{1}{r}\,H(\Theta),\qquad H(\Theta) := P_{\Theta^\perp}\nabla_\Theta\ell,\;}
\]
and taking Frobenius norms,
\[
  \Fro{\nabla_W L} = \frac{\Fro{H(\Theta)}}{\Fro{W}} \sim \frac{1}{\Fro{W}}.
\]
The direction-only quantity $\Fro{H(\Theta)}$ is $O(1)$ in $\Fro{W}$, so the raw gradient's Frobenius
norm shrinks like $1/\Fro{W}$: the larger $W$ grows, the smaller the raw gradient --- a natural brake.

\subsection{SGD vs Muon: who keeps the $1/\Fro{W}$ brake?}
\label{sec:sgd-vs-muon}

\subsubsection{SGD update scale}

SGD uses the raw gradient:
\[
  D_{\rm SGD} := W_{t+1} - W_t = -\eta\,G_t = -\frac{\eta}{r_t}H(\Theta_t),
\]
so
\[
  \Fro{D_{\rm SGD}} = \frac{\eta\,\Fro{H(\Theta_t)}}{\Fro{W_t}} = O\!\Big(\frac{\eta}{\Fro{W_t}}\Big).
\]
As $\Fro{W}$ grows, the SGD update shrinks like $1/\Fro{W}$.

\subsubsection{Muon update scale: msign annihilates the $1/\Fro{W}$ factor}

The RMS-matched Muon update (dropping momentum for clarity) is
\[
  D_{\rm Muon} := W_{t+1} - W_t = -\eta s\,P_t,\qquad
  P_t = \msign(G_t),\qquad s = 0.2\sqrt{\max(m,n)}.
\]
Here $\msign$ is defined through the SVD: for $M = U\Sigma V^\top$, $\msign(M) := U V^\top$. It is
scale-invariant, since $\alpha M = U(\alpha\Sigma)V^\top$ for $\alpha > 0$ has the same singular vectors:
\[
  \msign(\alpha M) = \msign(M),\qquad \alpha > 0.
\]
Because $G_t = H(\Theta_t)/r_t$ with $1/r_t > 0$ a positive scalar,
\[
  P_t = \msign(G_t) = \msign\!\Big(\frac{H(\Theta_t)}{r_t}\Big) = \msign\big(H(\Theta_t)\big).
\]
The $1/\Fro{W_t}$ factor is \emph{annihilated} by $\msign$. So $P_t$ is $O(1)$ in $\Fro{W_t}$
($P_t^\top P_t = V V^\top$ is a rank-$k$ projection, $\Fro{P_t}^2 = k$ with $k = \operatorname{rank}(G)$),
and
\[
  \Fro{D_{\rm Muon}} = \eta s\,\Fro{P_t} = \eta s\sqrt{k} = O(\eta).
\]
Comparing the two:
\[
  \Fro{D_{\rm SGD}} = O\!\Big(\frac{\eta}{\Fro{W}}\Big),\qquad
  \Fro{D_{\rm Muon}} = O(\eta).
\]
Muon has \emph{lost} the $1/\Fro{W}$ brake. This is the mechanism behind the claim ``Muon drifts the
norm faster than SGD'', made precise in Section~\ref{sec:finding1}.

\subsection{Finding 1: Muon grows the Frobenius / spectral norm faster than SGD}
\label{sec:finding1}

\subsubsection{Norm increment}

For any update $D$ (SGD or Muon), the change in squared Frobenius norm is
\[
  \Fro{W + D}^2 - \Fro{W}^2 = 2\langle W, D\rangle_F + \Fro{D}^2.
\]
Writing $R_t := \Fro{W_t}^2$,
\[
  \Delta R_t := R_{t+1} - R_t = 2\langle W_t, D_t\rangle_F + \Fro{D_t}^2,
\]
with two contributions:
\begin{itemize}\itemsep0pt
\item radial term $2\langle W_t, D_t\rangle_F$: how much the update pushes $W$ along itself, i.e.\ grows $\Fro{W}$;
\item update-energy term $\Fro{D_t}^2 \ge 0$: always non-negative.
\end{itemize}

\subsubsection{SGD: $\Fro{W_t} \sim t^{1/4}$}

Because $\langle W, G\rangle = 0$ (Corollary~B), the SGD radial term vanishes:
\[
  2\langle W_t, D_{\rm SGD}\rangle_F = -2\eta\langle W_t, G_t\rangle_F = 0,
\]
so
\[
  \Delta R_t^{\rm SGD} = \Fro{D_{\rm SGD}}^2 = \eta^2\Fro{G_t}^2.
\]
By Corollary~A, $\Fro{G_t}^2 = \Fro{H(\Theta_t)}^2/R_t = C^2/R_t$ (with $C = \Fro{H(\Theta_t)}$ an
$O(1)$ direction-only quantity), so
\[
  \Delta R_t^{\rm SGD} \sim \frac{\eta^2 C^2}{R_t}.
\]
As an ODE, $\dot R = a/R$ with $a := \eta^2 C^2$, i.e.\ $\tfrac12 \dot{(R^2)} = a$, giving
\[
  R(t)^2 = R(0)^2 + 2at \implies R(t) \sim \sqrt{2at}\quad(t\text{ large}),
\]
hence $\Fro{W_t}^{\rm SGD} = \sqrt{R(t)} \sim t^{1/4}$ (precisely $\sqrt{\eta C}\,t^{1/4}$).

\subsubsection{Muon: $\Fro{W_t} \sim t^{1/2}$}

For Muon, $D_t = -\eta s P_t$ with $P_t = \msign(G_t)$. As a baseline,
\[
  \langle W_t, D_t\rangle_F = -\eta s\langle W_t, P_t\rangle_F \approx 0
\]
(the sign matrix stays roughly orthogonal to $W$; see the empirical check below). Then
\[
  \Delta R_t^{\rm Muon} \approx \Fro{D_t}^2 = \eta^2 s^2\Fro{P_t}^2 = \eta^2 s^2 k_t,
\]
where $k_t = \operatorname{rank}(G_t) \le \min(m, n)$ is roughly a constant $k$, so
$\Delta R_t^{\rm Muon} \sim b$ with $b := \eta^2 s^2 k$. This is \emph{independent} of $R_t$, so
\[
  R(t) = R(0) + bt \sim t\quad(t\text{ large}),\qquad \Fro{W_t}^{\rm Muon} \sim t^{1/2}
\]
(precisely $\eta s\sqrt{k}\,t^{1/2}$). So under the same schedule Muon grows the Frobenius norm as
$t^{1/2}$ against SGD's $t^{1/4}$.

\subsubsection{Empirical check of scale invariance via $c_P$ and $c_G$}
\label{sec:finding1:cpcg}

Define the (cosine) alignments
\[
  c_{G,t} := \frac{\langle W_t, G_t\rangle_F}{\Fro{W_t}\Fro{G_t}},\qquad
  c_{P,t} := \frac{\langle W_t, P_t\rangle_F}{\Fro{W_t}\Fro{P_t}}.
\]
The theory predicts that if scale invariance holds then $c_{G,t} \approx 0$ (this is Corollary~B).
We measured $c_G$ and $c_P$ on a nanoGPT step-48K checkpoint over a fixed diagnostic batch, averaged
over 24 two-dimensional weights (6 layers $\times$ \{c\_attn, c\_proj, mlp.c\_fc, mlp.c\_proj\}):

\begin{center}
\small
\begin{tabular}{lcccc}
\toprule
Run @ step & $c_G$ & $|c_G|$ & $c_P$ & $|c_P|$ \\
\midrule
B0 (plain Muon) & $+0.0000$ & $0.0052$ & $-0.0007$ & $0.0104$ \\
M4plain (proj cap) & $-0.0003$ & $0.0033$ & $+0.0024$ & $0.0049$ \\
\bottomrule
\end{tabular}
\end{center}

The alignments are within a few times $10^{-3}$ of zero, consistent with approximate scale invariance
at this checkpoint. We stress that this is a single sanity check, not a validation of the assumption
across the whole trajectory.

\subsubsection{Spectral norm: Muon moves $\spec{W}$ more than SGD when $\Fro{W}$ is large}
\label{sec:finding1:spec}

Assume $\sigma_1(W)$ is simple, with top singular vectors $(u_1, v_1)$. First order,
\[
  \sigma_1(W + D) = \sigma_1(W) + u_1^\top D v_1 + O\!\Big(\frac{\Fro{D}^2}{\sigma_1 - \sigma_2}\Big).
\]
For SGD, $\Fro{D_{\rm SGD}} = O(\eta/\Fro{W})$, so $|u_1^\top D_{\rm SGD}v_1| = O(\eta/\Fro{W})$ and
(using $\sigma_1 = \Fro{W}\theta_1$ with $\theta_1 = O(1)$)
\[
  \frac{|\Delta\sigma_1^{\rm SGD}|}{\sigma_1} = O\!\Big(\frac{\eta}{\Fro{W}^2}\Big).
\]
For Muon, $\Fro{D_{\rm Muon}} = O(\eta)$, so $|u_1^\top D_{\rm Muon}v_1| = O(\eta)$ and
\[
  \frac{|\Delta\sigma_1^{\rm Muon}|}{\sigma_1} = O\!\Big(\frac{\eta}{\Fro{W}}\Big).
\]
Taking the ratio,
\[
  \frac{|\Delta\sigma_1^{\rm Muon}|/\sigma_1}{|\Delta\sigma_1^{\rm SGD}|/\sigma_1} \sim \Fro{W}.
\]
The larger $\Fro{W}$, the more Muon disturbs the spectral norm relative to SGD.

\subsection{Finding 2: after a cap, the spectral-norm second order is $\ge 0$}
\label{sec:finding2}

A spectral cap (at tolerance $\rho = 0$) removes the top component of an update $D_0$:
\[
  D = D_0 - \alpha\,u_1 v_1^\top,\qquad \alpha = u_1^\top D_0 v_1,
\]
so that $u_1^\top D v_1 = 0$. Then to first order
\[
  \sigma_1(W + D) = \sigma_1(W) + \underbrace{0}_{\text{cap kills 1st order}} + \text{(2nd order)} + \cdots
\]
Is the second-order term a problem (does it still push $\sigma_1$ up)? We show it is non-negative, so
the cap does not freeze the weight.

\subsubsection{Symmetric dilation}

For $W \in \mathbb{R}^{m\times n}$, form the symmetric dilation
\[
  \mathcal{H}(W) = \begin{pmatrix} 0 & W \\ W^\top & 0 \end{pmatrix} \in \mathbb{R}^{(m+n)\times(m+n)},
\]
whose eigenvalues are $\pm\sigma_1, \pm\sigma_2, \dots$ (padded with zeros), so
$\sigma_1(W) = \lambda_{\max}(\mathcal{H}(W))$. The standard second-order eigenvalue perturbation of a
symmetric matrix $A + E$,
\[
  \lambda_1(A + E) = \lambda_1 + z_1^\top E z_1 + \sum_{k\ne 1}\frac{(z_k^\top E z_1)^2}{\lambda_1 - \lambda_k} + \cdots,
\]
with $z_k$ the eigenvectors of $A$, applied to the dilation with
$a_j := u_j^\top D v_1$, $b_j := u_1^\top D v_j$ for $j > 1$, gives
\[
  \sigma_1(W + D) = \sigma_1 + u_1^\top D v_1
  + \sum_{j>1}\!\left[\frac{(a_j + b_j)^2}{4(\sigma_1 - \sigma_j)} + \frac{(a_j - b_j)^2}{4(\sigma_1 + \sigma_j)}\right] + O(\Fro{D}^3).
\]
Every second-order term is non-negative ($\sigma_1 - \sigma_j > 0$ is the spectral gap;
$\sigma_1 + \sigma_j > 0$; numerators are squares).

\subsubsection{Why the second order being $\ge 0$ is exactly what we want}

With the cap enforcing $u_1^\top D v_1 = 0$, the weight $W$ can still evolve --- and $K_Y$ can still
accumulate energy \emph{in non-top directions}, becoming more isotropic --- through three mechanisms:
\begin{enumerate}\itemsep0pt
\item \textbf{Non-top eigenvalue growth}: components of $D$ along $u_j v_j^\top$ ($j > 1$) directly
raise $\sigma_j$, spreading the energy;
\item \textbf{Top-vector rotation}: $a_j = u_j^\top D v_1 \ne 0$ or $b_j = u_1^\top D v_j \ne 0$ rotate
the top direction toward a new one through the second-order term;
\item \textbf{Top switching}: if some $\sigma_j$ grows past $\sigma_1$, the next step's cap acts on the
new top direction.
\end{enumerate}
So a spectral cap is not about freezing $W$: it converts ``keep pouring energy into the same top
direction'' into ``explore new directions in the orthogonal complement''. Since
$\sigma_1^2/\Fro{W}^2 = 1/\text{stable rank}$, capping the top while other directions grow raises the
stable rank, and $K_Y = W W^\top$ becomes more isotropic.

\subsection{Extension: the cap targets $H_\infty$ entropy; other entropies give other cap directions}
\label{sec:theory:entropy}

Write the squared-singular-value distribution of $W$ as
\[
  p_i = \frac{\sigma_i^2}{\sum_j \sigma_j^2},\qquad \sum_i p_i = 1,
\]
and the usual R\'enyi entropies
\[
  H_\infty = -\log p_{\max},\qquad
  H_2 = -\log\sum_i p_i^2,\qquad
  H_1 = -\sum_i p_i\log p_i.
\]

\textbf{The spectral cap $=$ $H_\infty$.} We have
\[
  e^{H_\infty} = \frac{1}{p_{\max}} = \frac{\Fro{W}^2}{\sigma_1^2} = \text{stable rank}(W),
\]
and
\[
  -dH_\infty = d(\log p_{\max}) = 2\,d\log\sigma_1 - 2\,d\log\Fro{W},
\]
so, keeping the dominant piece, $dH_\infty < 0$ iff $u_1^\top D v_1/\sigma_1 > \langle W, D\rangle/\Fro{W}^2$.
Enforcing $u_1^\top D v_1 \le 0$ (our cap) protects $H_\infty$.

\textbf{Participation rank $=$ $H_2$.} With
\[
  \mathrm{PR}(W) = e^{H_2} = \frac{1}{\sum_i p_i^2}
                = \frac{(\sum_i \sigma_i^2)^2}{\sum_i \sigma_i^4}
                = \frac{(\tr K_Y)^2}{\Fro{K_Y}^2},
\]
a more ``democratic'' measure of the number of effective directions, and
\[
  -dH_2 \propto \sum_i p_i^2\,d\log\sigma_i^2,
\]
the corresponding cap projection direction is $\sum_i p_i^2\,u_i v_i^\top$ --- it touches
\emph{all} directions $u_i v_i^\top$, weighted by $p_i^2$ (a multi-modal cap).

\textbf{Shannon $H_1$.} Similarly $-dH_1 = \sum_i \log p_i\,dp_i$, with direction
$\sum_i \log p_i\,u_i v_i^\top$.

We use the $H_\infty$ (spectral) cap because it is the cheapest:
\begin{enumerate}\itemsep0pt
\item the top pair $(u_1, v_1)$ is available from a few steps of power iteration;
\item it controls $\sigma_1$ directly (attention score $= \sigma_1$; MoE rank-1 collapse $= \sigma_1$);
\item $H_\infty$ collapse fires exactly when $\sigma_1$ dominates.
\end{enumerate}
Because the second order is $\ge 0$ (Section~\ref{sec:finding2}), capping only $\sigma_1$ lets the
remaining update pour energy into non-top directions, so $\tr K_Y$ grows while $\sigma_1$ is held ---
$p_1$ falls, $H_\infty$ rises indirectly, and $H_2$ and $H_1$ improve for free. All three case studies
use the $H_\infty$ cap.

\subsection{Remark: we do not actually need $K_X = I$}
\label{sec:theory:remark}

If $K_X \ne I$, with $K_Y = W K_X W^\top$, the ``spectral to control'' is not that of $W$ but of
\[
  A := W K_X^{1/2},\qquad K_Y = A A^\top,\qquad \sigma_i(A)^2 = \lambda_i(K_Y).
\]
So the object to cap is the spectral of $A$, per application:
\begin{itemize}\itemsep0pt
\item \textbf{nanoGPT FFN}: $K_X$ is the GELU-hidden second moment, anisotropic, so we cap $K_Y$
(a data-cap, not $W$);
\item \textbf{MoE router}: $K_X$ is the post-RMSNorm token covariance, relatively isotropic (RMS
removes the global scale);
\item \textbf{FlashAttention}: $K_X$ is the layer-input embedding covariance, anisotropic, so F3 caps
$K_Q^{(h)} = W_Q^{(h)\top} K_X W_Q^{(h)}$, not $W_Q$.
\end{itemize}
$K_X = I$ is only a pedagogical simplification (Muon still removes the brake, the cap still applies);
in practice the cap is on the data-coupled $K_Y$, not on the raw $W$.

\clearpage

\section{Application 1: nanoGPT FFN (proof of concept)}
\label{sec:nano}

\subsection{What $K_X$ is here}
\label{sec:nano:Kx}

\textbf{Module}: the \code{mlp.c\_proj} layer, which projects the GELU-activated hidden
($d_{\rm hidden} = 1536$) back into the residual stream ($d_{\rm embd} = 384$):
\[
  W \in \mathbb{R}^{d_{\rm embd}\times d_{\rm hidden}} = \mathbb{R}^{384\times 1536},
  \qquad X \in \mathbb{R}^{M\times d_{\rm hidden}}.
\]
$X$ is the GELU output, with $M \approx 32\times 1024 = 32768$ tokens per batch.

\textbf{$K_X$}: the second moment of the GELU-hidden activation,
$K_X = \tfrac{1}{M} X^\top X \in \mathbb{R}^{1536\times 1536}$, strongly anisotropic (after GELU only a
few channels are typically active).

\textbf{$K_Y$}: the covariance the FFN writes into the residual stream,
$K_Y = W K_X W^\top \in \mathbb{R}^{384\times 384}$. This is the geometric object the layer actually
contributes downstream, and the one we want isotropic.

\subsubsection{Estimating $K_Y$ and finding the top eigenpair $(q, \lambda_1)$}

At each Muon step:
\begin{enumerate}\itemsep0pt
\item take $X \in \mathbb{R}^{M\times 1536}$ from the forward-pass cache (sub-sampling $M = 4096$
tokens to avoid the memory cost of the full batch);
\item compute $Y = X W^\top \in \mathbb{R}^{M\times 384}$, cost $O(M\cdot d_{\rm embd}\cdot d_{\rm hidden})$;
\item do not form $K_Y$ explicitly (downstream only needs $\lambda_1$ and its $q$);
\item find the top eigenpair by \textbf{power iteration}: from a random $q_0 \in \mathbb{R}^{384}$,
iterate $q \leftarrow K_Y q / \Fro{K_Y q}$, with each $K_Y q = W K_X W^\top q$ done by matrix--vector
products (never forming $K_Y$).
\end{enumerate}

\textbf{Cost per step}: each power iteration is $O(M d_{\rm hidden} + M d_{\rm embd})$; with
$n_{\rm power} = 4$, total $\approx 4\times 4096\times 1920 \approx 31$ MFLOPs. Against the
$\sim$GFLOP forward+backward this is about $0.003\%$; the cap overhead is negligible.

\textbf{Cap first-order projection}, given $\lambda_1$ and $q$:
\[
  B = \frac{\partial\lambda_1}{\partial W} = 2qg^\top,\qquad
  g = \frac{1}{M}\sum_m (q^\top y_m)\,x_m,\qquad
  H \leftarrow H - \max\!\big(0, \langle B, H\rangle - \rho\lambda_1\big)\frac{B}{\Fro{B}^2}.
\]

\textbf{Geometry of $qg^\top$ (important)}: the rank-1 matrix $B = 2qg^\top$ is the direction in
weight space, at the current $W$, along which $\lambda_1(K_Y) = \lambda_1(W K_X W^\top)$ grows fastest.
Here $q$ is the most sensitive \emph{output-side} direction (top eigenvector of $K_Y$), $g$ is the
\emph{input-side} direction most aligned with $q$ (in the $K_X$ metric), and the outer product
$qg^\top$ couples them. Adding $B$ to $W$ raises $\lambda_1$ the fastest --- the direction we least
want $W$ to move along. The cap projects out the \emph{positive} part
$\langle B, H\rangle - \rho\lambda_1$ of the candidate update; the remainder lies in the orthogonal
complement of $B$, so $W$ still updates but no longer along the $\lambda_1$-growth direction.
$\rho = 0$ is a hard cap ($\lambda_1$ does not grow to first order); $\rho > 0$ leaves a buffer.

\subsection{Experimental setup}
\label{sec:nano:setup}

\begin{center}
\small
\begin{tabular}{ll}
\toprule
Setting & Value \\
\midrule
Model & nanoGPT 6L / 6H / $n_{\rm embd}=384$ / block 1024 \\
Optimiser & RMS-matched Muon (2D) + AdamW (norms / embed / lm\_head) \\
Precision & bf16 attention, fp32 weights \\
LR & $\times 10$ cosine decay $10^{-2} \to 10^{-3}$ \\
Iterations & 50K \\
WD & $0.0$ (removed, to test whether the cap works on its own) \\
Data & OpenWebText \\
Hardware & $2\times$ RTX 4090 \\
\bottomrule
\end{tabular}
\end{center}

Three setups:
\begin{itemize}\itemsep0pt
\item \textbf{D0}: no cap (baseline);
\item \textbf{D1}: W-cap on $W = $ \code{mlp.c\_proj} (directly controls $\sigma_1(W)$; corresponds
to the $K_X = I$ simplification);
\item \textbf{D1d}: data-cap on $K_Y$ (the actual $K_X \ne I$ case).
\end{itemize}

\subsection{Result: the three methods have equal val loss}
\label{sec:nano:results}

\begin{figure}[h]
\centering
\includegraphics[width=0.6\textwidth]{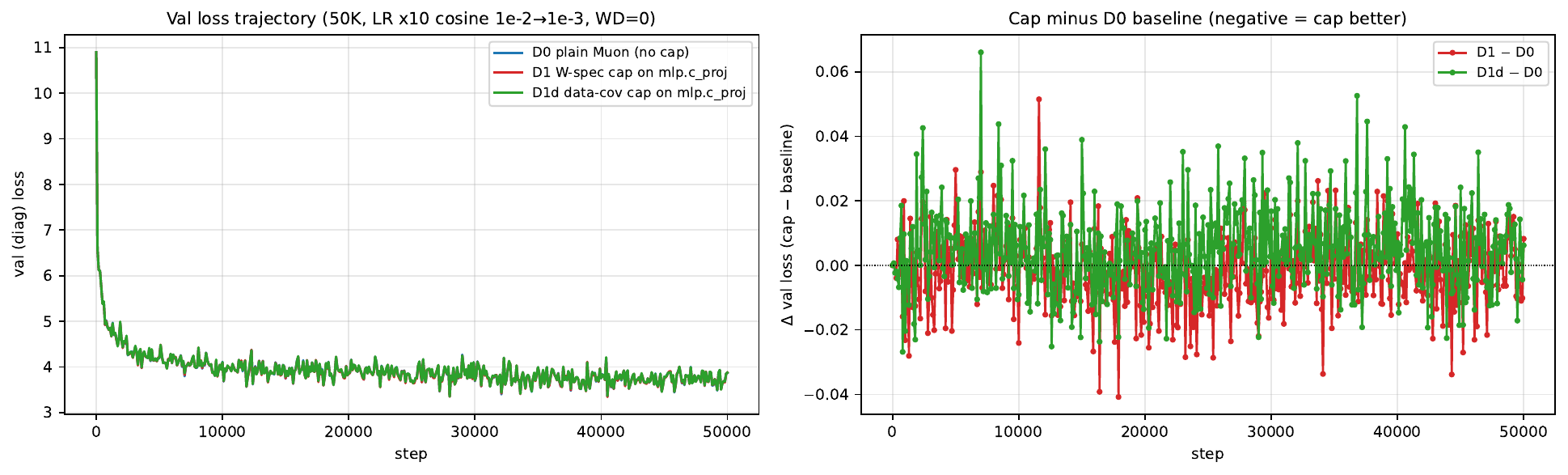}
\caption{nanoGPT 50K val-loss curves. D0/D1/D1d essentially overlap over the whole run, with a final
difference $|\Delta\text{val}| < 0.01$ nats. This is the proof of concept: the spectral cap
(projecting out first-order top-mode growth) does \emph{not} hurt val loss, so the model keeps
learning through the ``update learns new directions in the orthogonal complement'' mechanism of
Section~\ref{sec:finding2}.}
\label{fig:nano:loss}
\end{figure}

Final val loss @ 50K: D0 $= 3.8680$, D1 $= 3.8763$, D1d $= 3.8742$. The differences are $< 0.01$ nats,
well below the seed-to-seed noise ($\sim 0.05$).

\subsection{What the cap changes: spectral norm and isotropy side by side}
\label{sec:nano:tables}

\begin{center}
\small
\begin{minipage}{0.46\textwidth}
\centering
\textbf{Table A: per-layer $\sigma_1(W)$ @ 50K}\\[2pt]
\begin{tabular}{lccc}
\toprule
L & D0 & D1 (W-cap) & D1d (data-cap) \\
\midrule
L0 & 98.9  & \textbf{85.0}  & 100.6 \\
L1 & 163.3 & \textbf{116.6} & 165.0 \\
L2 & 137.4 & \textbf{117.2} & 139.6 \\
L3 & 156.4 & \textbf{128.5} & 157.8 \\
L4 & 219.4 & \textbf{117.4} & 164.5 \\
L5 & 211.2 & \textbf{145.2} & 191.6 \\
\bottomrule
\end{tabular}\\[2pt]
\footnotesize D1 lowers $\sigma_1$ by 14--46\%; D1d barely moves it.
\end{minipage}\hfill
\begin{minipage}{0.46\textwidth}
\centering
\textbf{Table B: per-layer $K_Y$ top-share @ 50K}\\[2pt]
\begin{tabular}{lccc}
\toprule
L & D0 & D1 (W-cap) & D1d (data-cap) \\
\midrule
L0 & 0.992 & 0.993 & \textbf{0.704} \\
L1 & 0.151 & 0.109 & \textbf{0.039} \\
L2 & 0.060 & 0.052 & 0.053 \\
L3 & 0.071 & 0.071 & 0.066 \\
L4 & 0.742 & 0.603 & \textbf{0.299} \\
L5 & 0.499 & 0.288 & 0.286 \\
\bottomrule
\end{tabular}\\[2pt]
\footnotesize D1d improves L1/L4 isotropy by 60--74\%.
\end{minipage}
\end{center}

\textbf{The two caps control different objects}: D1 presses $\sigma_1(W)$ directly (parameter space);
D1d presses $\lambda_1(K_Y) = \sigma_1(W K_X^{1/2})^2$ directly (output space). Because $K_X \ne I$
the two are not equivalent: D1 lowers the spectrum of $W$, but $K_X$ redistributes the energy onto the
strong data directions and the $K_Y$ top-share barely moves; D1d controls $K_Y$ directly while the
spectrum of $W$ can stay high, because $K_X^{1/2}$ orthogonalises it away.

This confirms the design choice at the end of Section~\ref{sec:theory:remark}: cap the data-coupled
$K_Y$ (no need to assume $K_X = I$). Capping only the spectrum of $W$ lets the strong directions of
$K_X$ keep the $K_Y$ top-share almost unchanged, and the cap fails to do its job.

\subsection{Participation-rank figure}
\label{sec:nano:pr}

\begin{figure}[h]
\centering
\includegraphics[width=0.85\textwidth]{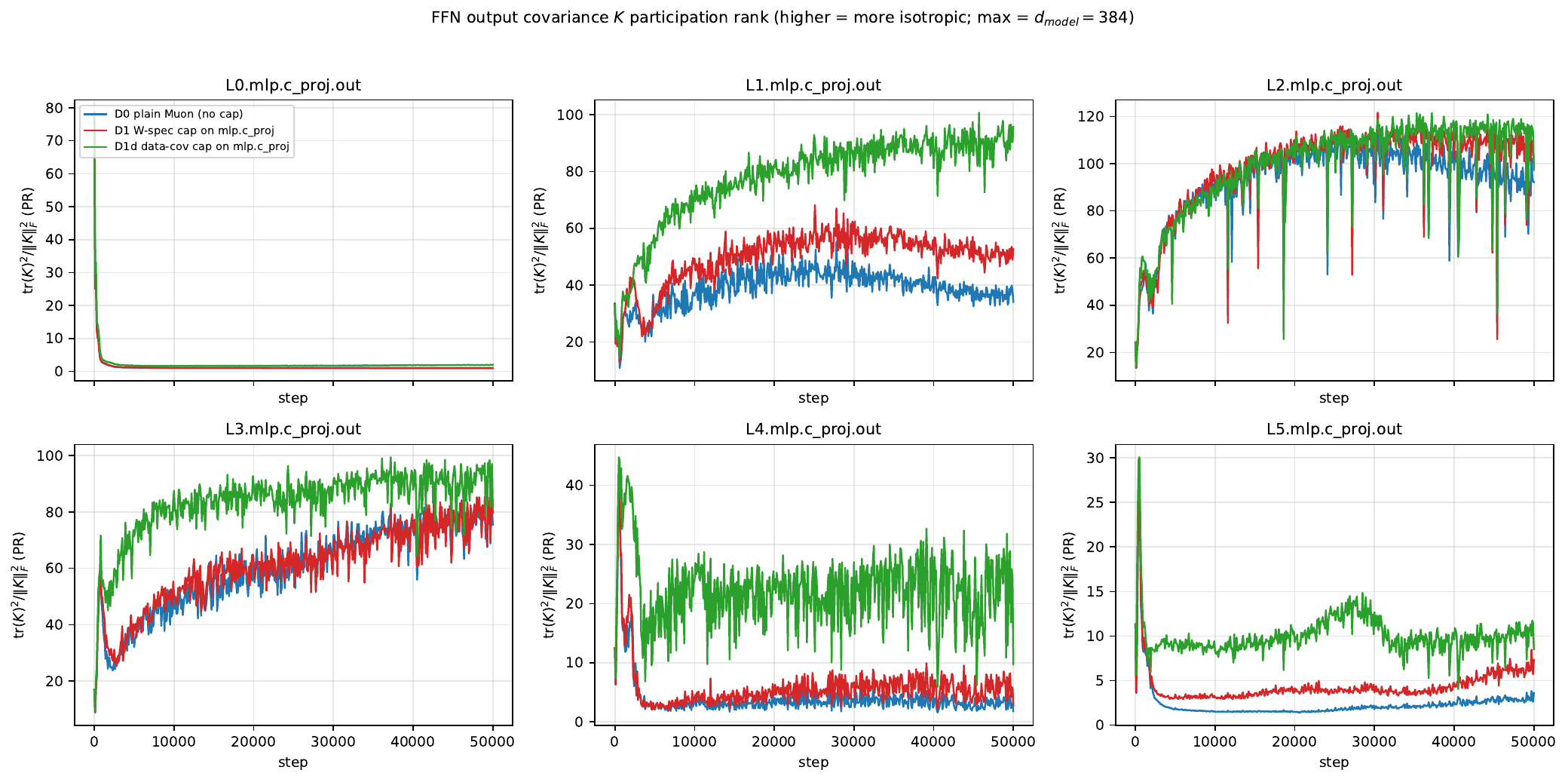}
\caption{Participation rank $\tr(K_Y)^2/\Fro{K_Y}^2$ (i.e.\ $e^{H_2}$ of
Section~\ref{sec:theory:entropy}, max $= 384$) of the FFN output $K_Y$ across the six
\code{mlp.c\_proj} layers. D1d (green) raises the L1 PR from D0's (blue) 34 to 96 (nearly $3\times$);
D1 (red) is in between. L0 is the hardest --- all three are near 1 (i.e.\ $K_Y$ is essentially rank-1
at L0).}
\label{fig:nano:pr}
\end{figure}

\textbf{Reading}: the rise in PR (i.e.\ higher $H_2$ entropy) is the core benefit of the D1d cap. The
cap stops the continued pouring into the $H_\infty$ direction ($\sigma_1$); the remaining update lets
other singular values grow, the energy distribution flattens, and $H_2$ improves automatically. This
is exactly the ``$H_\infty$ cap also improves $H_2$'' point of Section~\ref{sec:theory:entropy}.

\subsection{Part 2 summary}
\label{sec:nano:sum}

\begin{itemize}\itemsep0pt
\item the cap does not hurt val loss $\Rightarrow$ empirical support for the ``second order still
learns new directions'' mechanism of Section~\ref{sec:finding2};
\item the data-cap (D1d) is more on-target than the W-cap (D1) $\Rightarrow$ support for the ``cap
$K_Y$, do not assume $K_X = I$'' design of Section~\ref{sec:theory:remark};
\item PR rises from 34 to 96 $\Rightarrow$ support for ``$H_\infty$ cap improves $H_2$''.
\end{itemize}

nanoGPT's own val loss does not move, though; to see a real gain from the cap we turn to the MoE and
FA systems below.

\clearpage

\section{Application 2: 64-expert MoE router}
\label{sec:moe}

\subsection{What $K_X$ is here}
\label{sec:moe:Kx}

\textbf{Module}: the router of a MoE layer --- a linear layer mapping a token embedding to the logits
of $E = 64$ experts:
\[
  W_r \in \mathbb{R}^{E\times d} = \mathbb{R}^{64\times 384},\qquad X \in \mathbb{R}^{M\times d}.
\]
$X$ is the post-RMSNorm token embedding, $M$ the number of tokens sampled per step (the cap uses
$M = 4096$).

\textbf{$K_X$}: the covariance of the post-RMSNorm token embedding,
$K_X = \tfrac{1}{M}X^\top X \in \mathbb{R}^{384\times 384}$, relatively isotropic (RMS unifies the
overall scale).

\textbf{$K_z$ = expert-centred covariance of the raw router logits}:
\[
  Z = XW_r^\top \in \mathbb{R}^{M\times E},\quad
  Z_c = Z C,\quad
  C := I_E - \tfrac{1}{E}\mathbf{1}\mathbf{1}^\top,\quad
  K_z = \tfrac{1}{M}Z_c^\top Z_c \in \mathbb{R}^{64\times 64}.
\]
\textbf{Why the factor $C$}: softmax and top-$k$ routing are insensitive to a per-token global shift
(adding the same constant $c_t$ to all experts does not change the ordering). $C$ removes this gauge
freedom: $Z + c\mathbf{1}^\top \to (Z + c\mathbf{1}^\top)C = ZC$. So $K_z$ is the geometric quantity
that actually affects routing.

\textbf{$K_{z+b}$ = selection-score covariance}: once the loss-free bias (LFB) adds a per-expert bias
$b \in \mathbb{R}^E$, top-$k$ is taken on the selection score $s = z + b$, not on raw $z$. So the
geometry that truly decides routing is
\[
  K_{z+b} = \tfrac{1}{M}\big((Z + \mathbf{1}b^\top)C\big)^\top\big((Z + \mathbf{1}b^\top)C\big).
\]
$C$ removes only the per-token shift; $b$ is a per-expert shift, and $b \to b + c$ is also removed by
$C$ ($Cb$ has mean 0). So the direction the LFB actually learns, $Cb$, is retained in $K_{z+b}$.

\subsubsection{Algorithm: estimating $K_{z+b}$ and finding the top eigenpair}

At each Muon step:
\begin{enumerate}\itemsep0pt
\item take the cached $X \in \mathbb{R}^{M\times 384}$ ($M = 4096$ tokens);
\item compute $Z = XW_r^\top \in \mathbb{R}^{M\times 64}$, add $b$ to get $Z + b$, multiply by $C$ to
get $Z_c \in \mathbb{R}^{M\times 64}$;
\item form $K = Z_c^\top Z_c/M \in \mathbb{R}^{64\times 64}$ directly ($E = 64$ is small, so $O(M E^2)$
is affordable);
\item power iteration on $K$ for $(q, \lambda_1)$, four iterations at $O(E^2)$;
\item cap projection so that $\Delta\lambda_1 \le 0$ to first order.
\end{enumerate}
\textbf{Cost}: $O(M E^2 + M E d) \approx 100$ MFLOPs, again negligible against forward+backward.

\textbf{Three setups}:
\begin{itemize}\itemsep0pt
\item \textbf{G0}: LFB only (no cap);
\item \textbf{G1}: LFB + cap on the raw $K_z$ (without $b$);
\item \textbf{G3}: LFB + cap on $K_{z+b}$ (the selection-score geometry).
\end{itemize}

\subsection{What isotropy means in the MoE}
\label{sec:moe:meaning}

The top-share of $K_{z+b}$, $\text{top\_share} = \lambda_1/\tr K_{z+b}$, measures how much of the
variance of the selection score across tokens is concentrated in one direction:
\[
  \text{top\_share} \to \tfrac{1}{E-1} \approx 0.016 : \text{energy spread equally over expert directions};
\]
\[
  \text{top\_share} \to 1 : \text{every token's } s = z+b \text{ varies along a single fixed direction}.
\]

\textbf{The physics of top-share $\to 1$ (rank-1)}: each token $t$'s selection score is
\[
  s_t \approx c_t\cdot d_{\max} \in \mathbb{R}^E,
\]
with $d_{\max}$ a fixed direction and $c_t$ a token-dependent scalar. Then $\text{top-}k(s_t)$ is the
same as $\text{top-}k(d_{\max})$ --- all tokens route to the same $k$ experts. This is the
``one-expert dominates'' rank-1 collapse.

\textbf{Why the LFB cannot fix rank-1}: the LFB has only $E$ per-expert scalars, adjusting each
expert's threshold along the direction $d_{\max}$. If the geometry is one-dimensional, 63 of the 64
LFB degrees of freedom are wasted --- it can move an expert up or down but cannot make
\emph{different tokens pick different experts}.

\subsection{Core result: $K_{z+b}$ isotropy across the three methods}
\label{sec:moe:iso}

The three tables below are the final $K_{z+b}$ geometry (seed=42, fast MoE; G0v2/G1v2 are freshly run
baselines, G3 is the previously finished run at the same seed). The three metrics are complementary:
\begin{itemize}\itemsep0pt
\item \textbf{$\lambda_{\max}$}: absolute scale --- variance energy in the strongest direction;
\item \textbf{top-share} $= \lambda_{\max}/\tr K_{z+b}$: relative concentration (0 isotropic, 1 rank-1);
\item \textbf{participation rank} $= (\tr K_{z+b})^2/\Fro{K_{z+b}}^2$: effective number of directions
(upper bound $E-1 = 63$).
\end{itemize}

\begin{center}
\small
\begin{tabular}{lcccccc}
\toprule
$K_{z+b}$ \textbf{$\lambda_{\max}$} (smaller = more isotropic) & L0 & L1 & L2 & L3 & L4 & L5 \\
\midrule
\textbf{G0} (no cap) & \textbf{82{,}517} & \textbf{5{,}179} & \textbf{1{,}824} & \textbf{729} & \textbf{453} & \textbf{8{,}062} \\
\textbf{G1} (raw cap on $K_z$) & 2{,}052 & 2{,}173 & 66 & 71 & 121 & 7{,}142 \\
\textbf{G3} (sel cap on $K_{z+b}$) & \textbf{521} & \textbf{435} & \textbf{102} & \textbf{37} & \textbf{13} & \textbf{209} \\
\bottomrule
\end{tabular}
\end{center}

\textbf{Huge $\lambda_{\max}$ gaps}: G0 L0 $= 82{,}517$ is $158\times$ G3 L0 $= 521$; G0 L5 $= 8{,}062$
is $38\times$ G3 L5 $= 209$. The G1 raw cap leaves a clearly higher residual at L0/L5 than G3.

\begin{center}
\small
\begin{tabular}{lcccccc}
\toprule
$K_{z+b}$ \textbf{top-share} & L0 & L1 & L2 & L3 & L4 & L5 \\
\midrule
\textbf{G0} (no cap) & \textbf{0.995} & 0.667 & 0.653 & 0.422 & 0.356 & 0.864 \\
\textbf{G1} (raw cap on $K_z$) & 0.479 & 0.645 & 0.341 & 0.211 & 0.418 & 0.803 \\
\textbf{G3} (sel cap on $K_{z+b}$) & \textbf{0.272} & \textbf{0.521} & \textbf{0.305} & \textbf{0.119} & \textbf{0.173} & \textbf{0.267} \\
\bottomrule
\end{tabular}
\end{center}

\begin{center}
\small
\begin{tabular}{lcccccc}
\toprule
$K_{z+b}$ \textbf{participation rank} (max $= E-1 = 63$) & L0 & L1 & L2 & L3 & L4 & L5 \\
\midrule
G0 (no cap) & 1.0 & 2.1 & 2.3 & 4.8 & 5.1 & 1.3 \\
G1 (raw cap) & 3.2 & 2.2 & \textbf{6.6} & 11.2 & 4.8 & 1.5 \\
\textbf{G3} (sel cap) & \textbf{8.8} & \textbf{3.3} & 5.6 & \textbf{20.1} & \textbf{11.7} & \textbf{8.3} \\
\bottomrule
\end{tabular}
\end{center}

\textbf{Reading}:
\begin{enumerate}\itemsep0pt
\item \textbf{G0's L0 is a full rank-1 collapse}: top-share $= 0.995$, PR $= 1.0$ (exactly one
effective expert direction), $\lambda_{\max} = 82{,}517$ far above every other layer;
\item \textbf{G1 (raw cap on $K_z$) improves $K_{z+b}$ indirectly}: the raw cap controls $K_z$ (without
$b$), but once $K_z$ is held, $b$ grows in other directions to restore balance, so $K_{z+b}$ is also
better --- just not as good as G3;
\item \textbf{G3 (sel cap on $K_{z+b}$) wins on all three metrics}: across $6$ layers $\times 3$ metrics
$= 18$ cells, G3 takes 16; the other two (PR at L2 / L1) go marginally to G1;
\item \textbf{G3's L5 improvement is the largest}: from G0's top-share $0.86$ / $\lambda_{\max}$ $8062$
down to $0.27$ / $209$ ($38\times$). L5 has the weakest LFB feedback (last layer), so directly
protecting the geometry matters most there.
\end{enumerate}

\subsection{L0 / L5 load balancing: a phase transition and LFB failure}
\label{sec:moe:l05}

\textbf{Normalisation of the hard-load table}: write $F$ as the ``fraction of tokens that include
expert $e$ in their top-$k$'' --- summed over experts this is $k = 4$, with ideal uniform value
$k/E = 4/64 = 0.0625$. The table below has $F$ multiplied by $k = 4$ (i.e.\ sum-to-$k$ instead of
sum-to-1), so its maximum is $1.0$ (one expert appears in the top-4 of every token).

\begin{center}
\small
\begin{tabular}{lcccccc}
\toprule
\textbf{Effective experts per layer} (max $= 64$) & L0 & L1 & L2 & L3 & L4 & L5 \\
\midrule
G0 (no cap) & 31.3 & 64.0 & 64.0 & 64.0 & 63.9 & 31.0 \\
G1 (raw cap) & \textbf{62.2} & 51.7 & 63.9 & 63.9 & 63.9 & \textbf{14.3} \\
G3 (sel cap) & 46.3 & 63.9 & 63.9 & 64.0 & 63.9 & 30.1 \\
\bottomrule
\end{tabular}
\end{center}

\begin{center}
\small
\begin{tabular}{lcccccc}
\toprule
\textbf{Hard-load max} $F$ (uniform $= 0.0625$; rank-1 $\to 1.0$) & L0 & L1 & L2 & L3 & L4 & L5 \\
\midrule
G0 (no cap) & \textbf{0.999} & 0.066 & 0.069 & 0.069 & 0.072 & 0.820 \\
G1 (raw cap) & \textbf{0.071} & \textbf{0.139} & 0.071 & 0.070 & 0.074 & \textbf{0.894} \\
G3 (sel cap) & 0.219 & 0.075 & 0.072 & 0.067 & 0.075 & \textbf{0.236} \\
\bottomrule
\end{tabular}
\end{center}

\subsubsection{The L0 story: G0 collapses to rank-1}

G0 L0 hard-max $= 0.999$ --- literally the same expert appears in the top-4 of every token
(the hard-count signature of a first-order rank-1 collapse). effE $= 31.3$ reflects that the other 3
top-4 slots are still spread over many experts, but the top-1 expert is fixed.

G1 raw cap: hard-max $= 0.071 \approx$ uniform 0.0625, a clean rescue of the L0 rank-1 collapse.
effE $= 62.2$ is near the ceiling of 64.

G3 sel cap: hard-max $= 0.219$, in between --- not as clean as G1 (partial recovery), but far better
than G0's 0.999. effE $= 46.3$.

\subsubsection{The L5 story: G3 sel cap wins}

L5 hard-max:
\begin{itemize}\itemsep0pt
\item G0: 0.820 (the top expert takes 82\% of the tokens' top-4);
\item G1: \textbf{0.894} (worse --- the G1 raw cap has a side effect at L5);
\item \textbf{G3: 0.236} ($3$--$4\times$ better than G0/G1, close to uniform 0.0625).
\end{itemize}
This matches the $K_{z+b}$ top-share ordering of Section~\ref{sec:moe:iso} exactly: G3 breaks the L5
geometry, and the hard-load follows.

\subsection{Val loss and a late-decay reversal}
\label{sec:moe:val}

\begin{center}
\small
\begin{tabular}{lc}
\toprule
Run @ step 50K (seed=42 fast) & val loss \\
\midrule
G0 (no cap) & 3.7351 \\
\textbf{G1 (raw cap)} & \textbf{3.6660} \\
G3 (sel cap) & 3.6920 \\
\bottomrule
\end{tabular}
\end{center}

The three final val losses are within 0.07 nats.

\begin{figure}[h]
\centering
\includegraphics[width=0.75\textwidth]{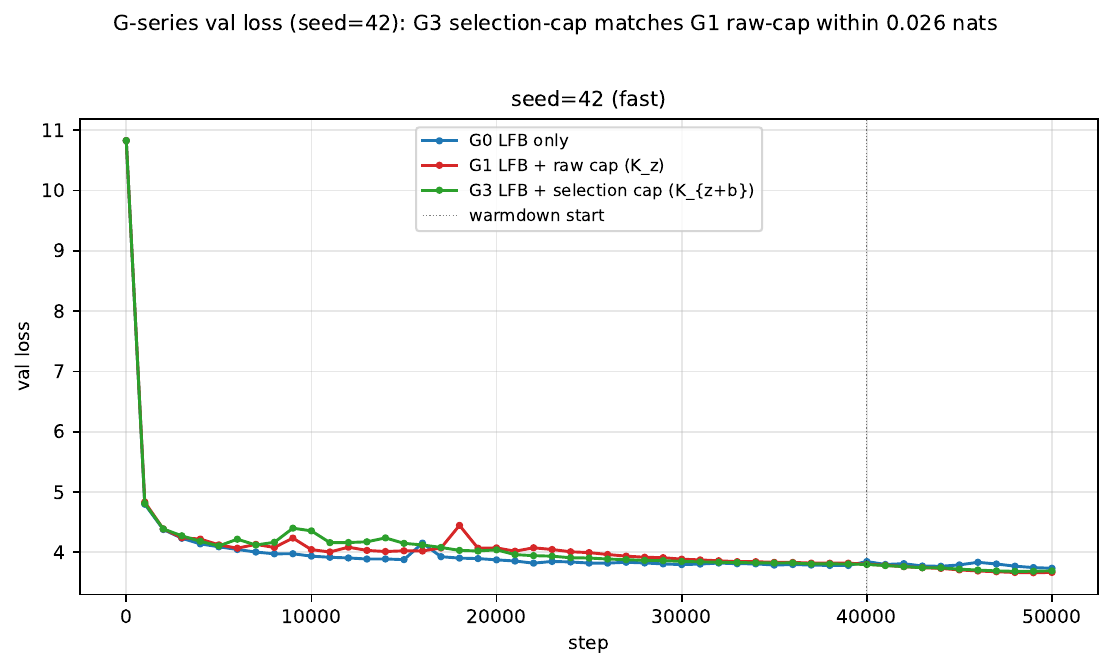}
\caption{seed=42 val-loss trajectory, three methods overlaid. \textbf{Key observation}: during
step 10K--30K, G0's val is slightly below G1/G3 (the cap constrains the router's directional freedom);
during the lr decay at step 40K--50K (warmdown start marked by the dashed line), the capped methods
G1/G3 overtake G0. This is the ``slower early, more stable late'' signature of the cap: it prevents
G0's fast early loss drop bought with over-anisotropic geometry, but G0's later spectral collapse
worsens, and the capped methods are more stable through the warmdown.}
\label{fig:moe:val}
\end{figure}

\subsection{Load-balance visualisation}
\label{sec:moe:vis}

\begin{figure}[htbp]
\centering
\includegraphics[width=\textwidth]{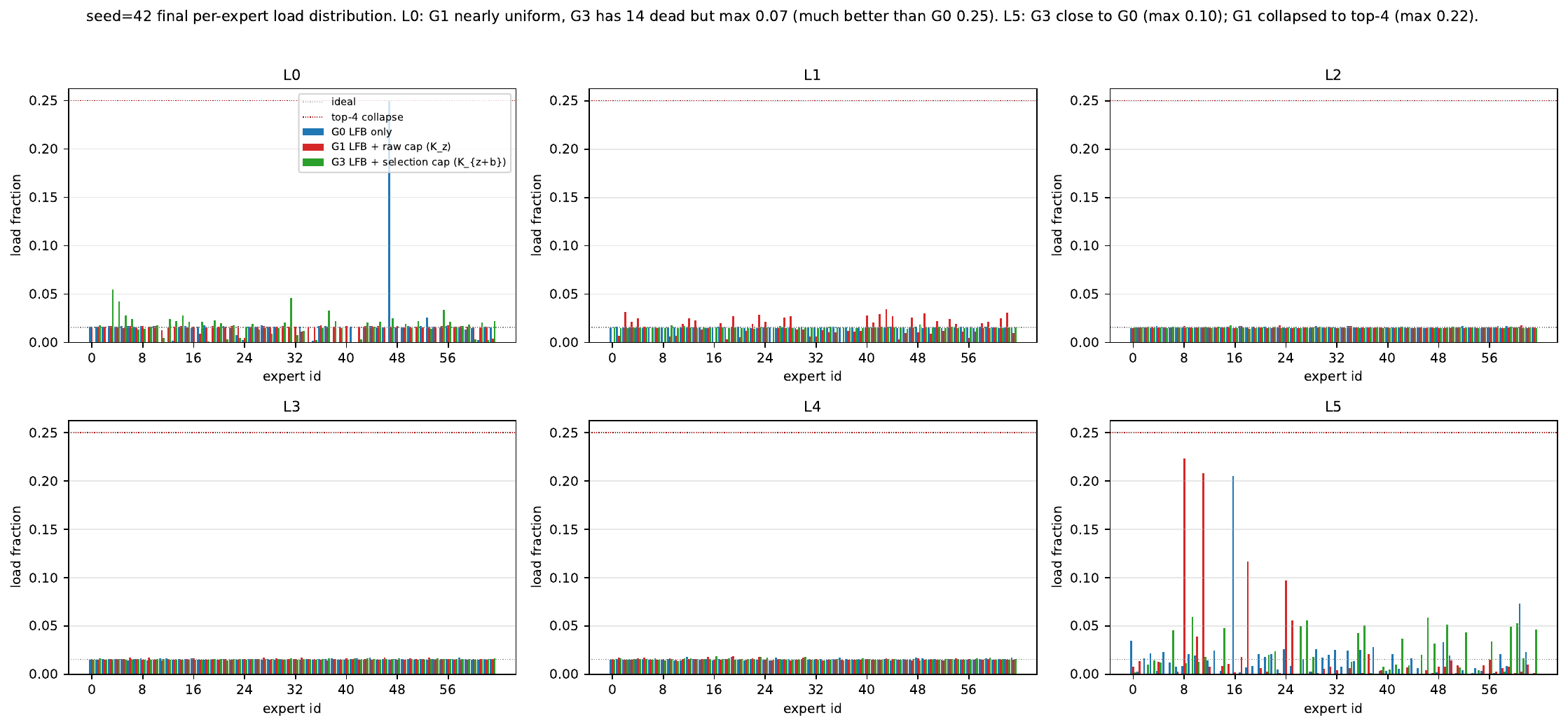}
\caption{seed=42 final per-expert load histogram (3 methods $\times$ 6 layers). In each subplot the
horizontal axis is the 64 experts (sorted by load), the vertical axis the load fraction, with the
ideal $1/E = 0.0156$ as a dashed line. \textbf{L0 (top row)}: G0 (blue) has its first expert take 25\%
(the rank-1 collapse, visualised); G1 and G3 both flatten to near-ideal. L1--L4: the three methods are
essentially the same. \textbf{L5 (bottom row)}: G1's top 4 experts take 65\% (a hard collapse); G3
repairs the G1 L5 side effect.}
\label{fig:moe:bar}
\end{figure}

\begin{figure}[h]
\centering
\includegraphics[width=0.85\textwidth]{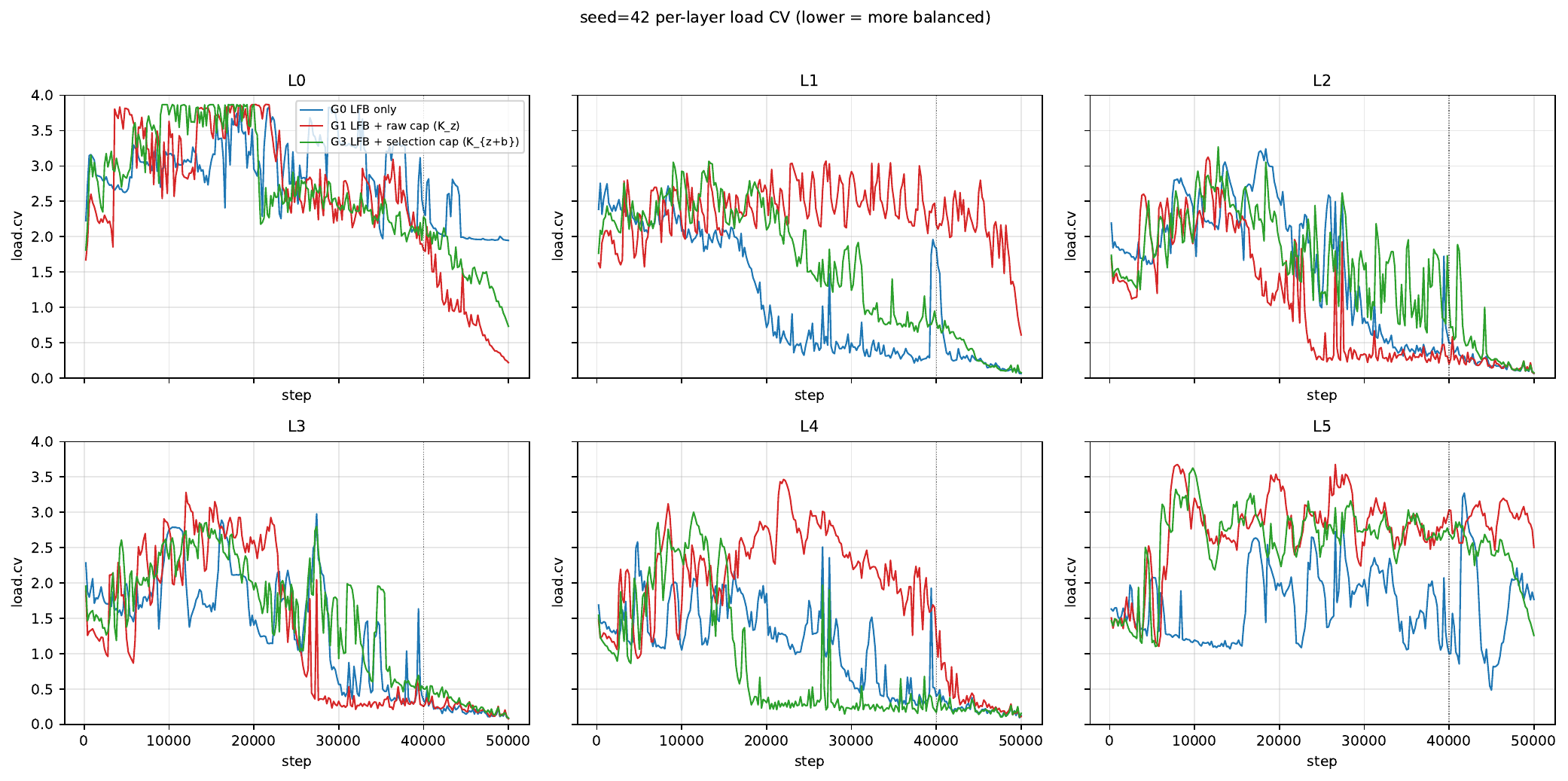}
\caption{Per-layer load CV ($E\cdot\text{std}(F)/\text{mean}(F)$, lower is more balanced, ideal $= 0$).
L0: G0 $= 1.96$ (very imbalanced), G1 $= 0.30$ (near perfect), G3 $= 0.92$ (in between). L1 is the
reverse: G1 $= 1.07$ (side effect), G3 $= 0.07$ (repaired). L5: G3 $= 1.50 <$ G0 $= 1.88 <$ G1
$= 2.86$.}
\label{fig:moe:cv}
\end{figure}

\subsection{G0's spectral norm is large and very anisotropic, yet most layers balance --- why?}
\label{sec:moe:discuss}

Back to G0's $K_{z+b}$ table: L1--L4 have top-share $0.35$--$0.67$ and $\lambda_{\max}$ from hundreds
to thousands --- still quite anisotropic (far from the ideal top-share $\approx 0.016$); yet their
hard-max is all $0.066$--$0.072$ (near uniform 0.0625).

\textbf{Explanation} (with Section~\ref{sec:moe:meaning}):
\begin{enumerate}\itemsep0pt
\item Muon injects a fixed-energy update into $W_r$ each step; without a cap, $\lambda_1(K_{z+b})$
keeps rising, but the LFB uses $b$ to ``pull the direction back to balance'';
\item in the ``moderately anisotropic'' regime top-share $\in [0.35, 0.7]$, the LFB's 64 degrees of
freedom are \emph{enough} to level the hard count --- it shifts each expert's selection threshold so
that different tokens pick different experts;
\item \textbf{but} the phase transition to top-share $\to 1$ (rank-1 collapse) breaks the LFB: at
rank-1, all tokens' scores are scalar multiples of a single direction, and a threshold shift cannot
make \emph{different tokens pick different experts};
\item G0 L0 / L5 sit right at this transition (top-share $0.86$--$0.99$), so their hard-max explodes;
\item G1 / G3 push all layers below the transition through the cap, leaving room for the LFB.
\end{enumerate}

\textbf{This also explains why most of G0's layers look fine}: the cap is not a \emph{necessary}
condition for balance, but a \emph{preventative} against the rank-1 phase transition. Its value is in
the worst case (L0 / L5), not the typical case (L1--L4).

\subsection{Part 3 summary}
\label{sec:moe:sum}

\begin{itemize}\itemsep0pt
\item the selection cap (G3) wins almost across the board on $K_{z+b}$ isotropy over all six layers;
\item G0's L0 enters the rank-1 phase transition at top-share $= 0.995$ $\Rightarrow$ the LFB fails
$\Rightarrow$ hard-max $\to 1.0$; the G1/G3 caps repair it (G1 cleanly, $= 0.071$; G3 partially,
$= 0.219$);
\item during lr decay the capped methods overtake the baseline, possibly because a healthier geometry
finds good directions more easily at small LR;
\item at L5 the cap is not always perfect on the hard count, but it does improve the geometry.
\end{itemize}

\clearpage

\section{Application 3: FlashAttention bf16 + L1 H2 rescue}
\label{sec:fa}

\subsection{What $K_X$ is here}
\label{sec:fa:Kx}

\textbf{Module}: the Q and K projections of each attention block. Each layer has 12 heads, each
projecting the $d_{\rm embd} = 768$ input embedding into a $d_h = 64$ Q/K subspace:
\[
  W_Q^{(h)},\,W_K^{(h)} \in \mathbb{R}^{d_h\times d_{\rm embd}} = \mathbb{R}^{64\times 768}.
\]

\textbf{$K_X$}: the input-embedding covariance at layer $\ell$,
$K_X^{(\ell)} = \tfrac{1}{M}X_\ell^\top X_\ell \in \mathbb{R}^{768\times 768}$, where $X_\ell$ is the
residual-stream token embedding at the entry of layer $\ell$. Strongly anisotropic (a typical
transformer input space has a few dominant directions).

\textbf{Cap target $=$ per-(layer, head) data covariance}:
\[
  K_Q^{(\ell, h)} = W_Q^{(\ell, h)} K_X^{(\ell)} W_Q^{(\ell, h)\top}
                  = \tfrac{1}{M}(Q^{(\ell, h)})^\top Q^{(\ell, h)} \in \mathbb{R}^{64\times 64},
\]
with $Q^{(\ell, h)} = X_\ell W_Q^{(\ell, h)\top}$ the query matrix of that head on the current batch;
the K side is symmetric.

\textbf{Why per-(layer, head)}: attention failure is a \emph{local} loss of control at one head in one
layer. A layer-aggregate cap reacts too slowly to a single head's phase change, so we run the cap
separately on each (layer, head).

\subsubsection{Algorithm and cost}

At each Muon step:
\begin{enumerate}\itemsep0pt
\item take $Q^{(\ell, h)} \in \mathbb{R}^{M\times 64}$ per head from the forward-pass cache
($M = 4096$ tokens per rank);
\item form $K_Q^{(\ell, h)} = (Q^{(\ell, h)})^\top Q^{(\ell, h)}/M \in \mathbb{R}^{64\times 64}$,
cost $O(M d_h^2) \approx 17$ MFLOPs per head;
\item power iteration on $K_Q^{(\ell, h)}$ for $(q, \lambda_1)$, four iterations at $O(d_h^2)$;
\item cap projection so that $\Delta\lambda_1 \le 0$; repeat on the K side.
\end{enumerate}

\textbf{Total cost} (12 layers $\times$ 12 heads $\times$ 2 (Q+K) $= 288$ caps per step): $\sim 5$
GFLOPs, about 5\% of the $\sim 100$ GFLOP forward+backward --- still acceptable.

\subsection{The bf16 FA failure mechanism (from the reference paper)}
\label{sec:fa:paper}

The reference paper on low-precision FlashAttention failures~\cite{lp2510} identifies the mechanism:
FlashAttention v2 uses \emph{block-wise online softmax}. For each query row the score matrix is split
into blocks, each block's max and log-sum-exp are computed, and the blocks are merged. With a bf16
mantissa of 7 bits, at a score scale of $\sim 10^4$ the resolution is only $\sim 80$.

\textbf{Failure chain}:
\begin{enumerate}\itemsep0pt
\item late in training, some layer's $W_Q^{(\ell, h)}$ and $W_K^{(\ell, h)}$ accumulate energy along
the strong directions of $K_X^{(\ell)}$;
\item $\lambda_{\max}(K_Q^{(\ell, h)}) = \sigma_1(W_Q^{(\ell, h)}K_X^{(\ell)1/2})^2$ blows up;
\item the attention score $s_{ij} = q_i\cdot k_j/\sqrt{d_h}$, bounded by
$|s_{ij}| \le \sqrt{\lambda_Q\lambda_K}/\sqrt{d_h}$, grows with it;
\item when a row has \emph{many} entries near its row max (large near-max count), the cross-block
max-subtraction loses precision in bf16;
\item the softmax accumulates error into the attention output;
\item loss creep, then crash.
\end{enumerate}

\textbf{Timeline of the uncapped baseline}:
\begin{itemize}\itemsep0pt
\item iter 70K: onset, loss $= 3.33$ (creep begins);
\item iter 100K: late creep, loss $= 3.76$;
\item iter 110K: crash, loss $= 4.31$;
\item iter 120K: blown, loss $= 5.19$.
\end{itemize}

The paper fixes this with a patched (STABLE) softmax that adds a numerical safeguard inside attention.
\textbf{Our F3 leaves softmax untouched} and instead uses the cap to stop
$\lambda_{\max}(K_Q^{(\ell, h)})$ from reaching the failure threshold.

\textbf{Why the cap should work}: it guarantees $\Delta\lambda_{\max} \le 0$ per step to first order.
The crash path needs $\lambda$ to grow from $\sim 10^3$ to $\sim 10^5$ over 20K steps, i.e.\ an average
$\Delta\lambda \approx 5$ per step. Additionally, because the largest direction is capped each step,
the matrix explores other directions rather than pouring energy into the same direction --- which also
avoids producing many equal scores.

\subsection{F3 experimental setup}
\label{sec:fa:setup}

\begin{center}
\small
\begin{tabular}{ll}
\toprule
Setting & Value \\
\midrule
Model & GPT-2 small (12L / 12H / $n_{\rm embd}=768$ / block 1024) \\
Optimiser & RMS-matched Muon (2D) + AdamW \\
Precision & bf16 attention (\code{STABLE=0}, the paper's original unstable softmax, \textbf{no patch}) \\
LR & 5e-4, WD $= 0$, warmup 2000 \\
Batch & 327{,}680 tok/iter \\
Cap & \code{CAP\_MODE=qk\_data}, per-(layer, head) $\rho = 0$ hard cap \\
Iterations & 120K \\
Monitoring & \code{diag\_attn\_layers=1,11} (L1 and L11 record fine-grained score / near-max; all 12 layers are capped and log $\lambda_{\max}$) \\
\bottomrule
\end{tabular}
\end{center}

\textbf{Comparison}: the same configuration without a cap --- the uncapped baseline run.

\subsection{The full L1 H2 crisis and rescue}
\label{sec:fa:L1H2}

\textbf{Indexing note}: the paper uses 1-indexed layers, so the paper's ``L2'' is our L1 (the second
layer). Among the 24 monitored (layer, head) pairs, \textbf{L1 H2} (the third head of the paper's L2)
has the largest \code{score.max} peak, 27{,}264 at iter 86K. It corresponds to the bf16 FA failure
entry the paper observes at its L2. This subsection tracks \emph{this one head} throughout.

\subsubsection{Precise definitions of the diagnostic metrics (read before the table)}
\label{sec:fa:L1H2:defs}

The diagnostics are computed on one captured forward pass (\code{diag\_capture=True}). The attention
score tensor is
\[
  S \in \mathbb{R}^{B\times H\times T_q\times T_k},\qquad
  S_{b,h,i,j} = \frac{q_{b,h,i}\cdot k_{b,h,j}}{\sqrt{d_h}},
\]
with $B$ the diagnostic batch size, $H = 12$ heads, $T_q = T_k = 1024$, $d_h = 64$. The metrics below
are all defined \emph{for a fixed head} $h$ (here $h = 2$, i.e.\ L1 H2). The causal mask sets
$j > i$ to $-\infty$, and only finite entries are counted in the sums/maxima.

\textbf{(1) score.max (per head, the single largest entry)}:
\[
  \text{score.max}^{(h)} := \max_{b, i, j}\;S_{b, h, i, j}.
\]
Code: \code{S.amax(dim=(0,2,3))}. This is the single largest entry over all (batch, query, key)
triples, not the average of the row maxima. Physically: the worst-case bf16 stress (the entry most
squeezed by the bf16 mantissa).

\textbf{(2) Row-tie indicator + near-max count}: first the per-query row max
$\text{rowmax}^{(h)}_{b, i} := \max_j S_{b, h, i, j}$, then the per-query row-tie count (how many
entries in a row are within a $10^{-3}$ tolerance of the row max),
\[
  n^{(h)}_{b, i} := \big|\{ j : S_{b,h,i,j} \ge \text{rowmax}^{(h)}_{b, i} - 10^{-3}\}\big|.
\]

\textbf{(3) near.mean (per head, the mean row-tie count)}:
\[
  \text{near.mean}^{(h)} := \frac{1}{B T_q}\sum_{b, i}\;n^{(h)}_{b, i}.
\]
Healthy attention (one dominant softmax peak) gives $\approx 1.0$. A large value means most rows have
several ties, so the paper's biased-rounding mechanism fires \emph{frequently}.

\textbf{(4) near.max (per head, the worst row's tie count)}:
\[
  \text{near.max}^{(h)} := \max_{b, i}\;n^{(h)}_{b, i}.
\]
Value $K$ means at least one query has $K$ entries all within row max $\pm 10^{-3}$ --- the worst-case
trigger strength.

\textbf{(5) $\lambda_{\max}(K_Q^{(h)})$ and q\_excess}: see Section~\ref{sec:fa:Kx} and
Section~\ref{sec:fa:excess}. q\_excess is the cap's \code{excess\_mean} on the Q side at that step
(already including LR; units $=$ first-order change of $\lambda_1$ per Muon step).

\textbf{(6) top1 prob} $= \langle\max_j P_{b, h, i, j}\rangle_{b, i}$ with $P = \text{softmax}(S)$: the
mean top probability per row. High $=$ peaked attention.

\subsubsection{The main L1 H2 diagnostic table (key iterations)}
\label{sec:fa:L1H2:table}

All numbers are \textbf{specifically for L1 H2}, not layer-aggregate.

\begin{center}
\scriptsize
\begin{tabular}{rrrrrrrrr}
\toprule
iter & $\lambda_{\max}(K_Q^{L1,H2})$ & $\lambda_{\max}(K_K^{L1,H2})$ & q\_excess & k\_excess & score.max & near.mean & near.max & top1 prob \\
\midrule
200    & 3.2   & 3.2     & 0.000 & 0.000 & 2      & 1.02 & 4  & 0.012 \\
5K     & 13.2  & 8.0     & 0.021 & 0.011 & 7      & 1.02 & 3  & 0.057 \\
30K    & 189.9 & 156.9   & 0.098 & 0.087 & 64     & 1.53 & 11 & 0.114 \\
60K    & 502.4 & 872.8   & 0.133 & 0.130 & 209    & 1.91 & 12 & 0.246 \\
\textbf{70K} & \textbf{4{,}527} & \textbf{11{,}435} & \textbf{5.000} & \textbf{4.330} & \textbf{1{,}144} & \textbf{6.54} & \textbf{32} & 0.334 \\
75K    & 57{,}088 & 97{,}092 & 49.31 & 42.39 & 13{,}568 & 2.60 & 15 & 0.632 \\
80K    & 61{,}633 & 124{,}017 & 27.89 & 28.46 & 24{,}064 & 1.60 & 9 & 0.781 \\
\textbf{86K} & \textbf{101{,}698} & \textbf{133{,}208} \emph{peak} & 22.39 & 15.95 & \textbf{27{,}264} \emph{peak} & 1.39 & 7 & 0.845 \\
90K    & 93{,}315 & 129{,}284 & 17.47 & 17.71 & 25{,}600 & 1.38 & 8 & 0.845 \\
100K   & 31{,}657 & 76{,}344 & 12.49 & 9.75 & 11{,}136 & 1.23 & 6 & 0.901 \\
120K   & \textbf{1{,}008} & \textbf{31{,}741} & \textbf{0.42} & \textbf{0.17} & \textbf{2{,}272} & 1.16 & 5 & 0.921 \\
\bottomrule
\end{tabular}
\end{center}

Column definitions are in Section~\ref{sec:fa:L1H2:defs}. In one line: \code{score.max} is the single
global max entry (not an average), \code{near.mean} the mean per-row tie count, \code{near.max} the
worst-row tie count, and \code{q\_excess} already includes LR and is on the same scale as $\lambda_1$.

\subsubsection{The physical events, stage by stage (L1 H2 only)}

\textbf{Stage A (0--60K)}: a healthy plateau. $\lambda$ grows slowly from 3 to 873; score.max from 2
to 209; near.mean $= 1.91$ (normal attention is $\approx 1$). The cap is already trimming but the
excess is small ($0.13$).

\textbf{Stage B (60K--70K)}: entering the danger zone. In 10K steps $\lambda(K_K) = 873 \to 11{,}435$
($13\times$), score $= 209 \to 1144$ ($5\times$), and \textbf{near.mean rises from 1.91 to 6.54}
($3.4\times$). This is exactly the entry into the uncapped baseline's creep.

\textbf{Stage C (70K--86K)}: the cap enters crisis mode. The cap excess escalates:
\[
  60K: 0.13\;\to\; 70K: 5.0\;\to\; 75K: 49\;\text{(peak crisis)}\;\to\; 86K: 22.
\]
\textbf{The meaning of q\_excess $= 49$} (see Section~\ref{sec:fa:excess}): at that step the cap trims a
first-order $\Delta\lambda$-growth direction of magnitude 49 nats out of the Muon update, i.e.\ \emph{if
uncapped, the next step's $\lambda_{\max}$ would rise by 49}. But the cap is a first-order projection
and the second order can still grow $\lambda$: from 70K to 86K, $\lambda$ actually rises from 11K to
133K (the cap presses each step's $\Delta\lambda$ from an uncapped 50+ down to $\sim 10$, so $\lambda$
grows about $5\times$ more slowly).

\textbf{Stage D (86K, peak)}: the crisis peak. $\lambda_K = 133{,}208$, score $= 27{,}264$ (about half
the bf16 max of 65504). \textbf{But near.mean $= 1.39$, near.max $= 7$} (both down sharply).

\emph{Why near-max falls: attention becomes peaked}. At 70K the attention top1 prob is $0.33$ (spread);
at 86K it is $0.85$ (very peaked). Spread attention with several near-equal entries per row gives a
high near-max; peaked attention with one dominant entry gives a low near-max. This is not designed into
the cap --- it is a by-product of the transformer's own learning dynamics; but the cap itself avoids an
exponential blow-up while encouraging continued learning in new directions, so training settles into a
stable direction (see the figures) in which most tokens' attention is highly concentrated.

\textbf{Stage E (86K--120K)}: monotone cool-down. The cap keeps redirecting $W$ away from the
high-$\lambda$ basin. $\lambda_K = 133K \to 76K \to 32K$, score $= 27K \to 11K \to 2K$. The system
fully recovers; by 120K the cap excess is $= 0.17$, below the level at step 200.

\subsubsection{Counterfactual: the uncapped baseline over the same window}

\begin{center}
\small
\begin{tabular}{rll}
\toprule
iter & F3 (with cap) & uncapped baseline \\
\midrule
60K  & loss $= 3.13$, $\lambda_K^{L1H2} = 873$ & loss $= 3.03$, healthy \\
70K  & loss $= 3.11$, $\lambda_K^{L1H2} = 11.4$K \emph{cap crisis enters} & loss $= 3.33$, onset \\
80K  & loss $= 3.04$, near.max $= 9$ (bf16 critical) & loss $= 3.50$, creep \\
86K  & loss $= 3.00$, $\lambda$ peak 133K, cap trims 22 nats/step & loss $\sim 3.55$ \\
100K & loss $= 3.11$, $\lambda$ already down to 76K & loss $= 3.76$, late creep \\
\textbf{110K} & \textbf{loss $= 3.07$} (still stable) & \textbf{loss $= 4.31$, crash} \\
120K & loss $= 2.94$, $\lambda = 32$K, fully recovered & loss $= 5.18$ \\
150K & (not run) & loss $= 7.63$, blown \\
\bottomrule
\end{tabular}
\end{center}

F3's loss stays stable through the entire crash window (70K--110K) while the uncapped baseline crashes.
This is the strongest direct evidence of the cap's value.

\subsection{The physical meaning of \code{excess\_mean} (the cap's counterfactual metric)}
\label{sec:fa:excess}

Definition:
\[
  \text{excess\_mean}_t^{(\ell, h, \mathrm{Q})} := \max\!\Big(0,\;\langle B_t^{(\ell, h)}, H_t^{(\ell, h)}\rangle_F - \rho\lambda_1^{(\ell, h)}\Big),
\]
where $B_t^{(\ell, h)} = \partial\lambda_1(K_Q^{(\ell, h)})/\partial W_Q^{(\ell, h)} = 2 q_1 g_1^\top$
is the $W_Q$ direction that grows $\lambda_1$ fastest; $\rho = 0$ is a hard cap.

\textbf{On the scale of $H$ (important)}: $H_t^{(\ell, h)}$ is the \emph{actual candidate update} of
that Muon step,
\[
  H = -\eta_{\rm Muon}^{\rm adj}\cdot P,\quad P = \msign(M),\quad \eta_{\rm Muon}^{\rm adj} = 0.2\sqrt{\max(m,n)}\cdot\eta_t
\]
$H$ \emph{already} carries the LR and the RMS-matched scale --- it is not the
raw $\msign$ direction nor an unscaled momentum. So \code{excess\_mean} is in units of the
\emph{change of $\lambda_1$ per actual Muon step}, not ``per unit LR''.

\textbf{Geometric reading}: first-order Taylor,
\[
  \lambda_1(W_Q + H) \approx \lambda_1(W_Q) + \langle B, H\rangle_F
\]
(no separate $\eta$ factor, since $H$ already includes it). Hence
\[
  \boxed{\;\text{excess\_mean} = X \;\Leftrightarrow\;
  \text{``if the cap did not act this step, the next step's $\lambda_1$ would rise by $X$ (LR included)''}.\;}
\]
The units match $\lambda_1$ itself (our $\lambda_{\max}(K_Q)$ ranges from $\sim 3$ early to $\sim 10^5$
at 86K; the excess is a scalar on the same scale).

\textbf{Verification on L1 H2}: at iter 86K, excess\_mean $= 22$. From 86K to 100K (14K steps), an
uncapped expected rise of $\sim 10$--$22$ per step (mean $\sim 15$) would accumulate a first-order
expected $\lambda$ increase of $\sim 200{,}000+$. Instead $\lambda$ \emph{falls} by 56K
($133K \to 76K$).

\subsection{Why our cap fixes what the paper's patch does not}
\label{sec:fa:why}

The paper fixes the problem with a patched (STABLE) softmax --- a numerical safeguard inside attention.
We use the cap to stop $\lambda_{\max}$ at the source, before it reaches the failure threshold. The two
act at different places:
\begin{itemize}\itemsep0pt
\item the paper changes attention-internal numerics $\Rightarrow$ even if $\lambda$ is large, softmax
does not blow up;
\item we change the Muon optimiser $\Rightarrow$ $\lambda$ never gets large enough to trouble softmax.
\end{itemize}

\textbf{Advantages of our approach}:
\begin{enumerate}\itemsep0pt
\item the cap does not touch attention internals, so it is compatible with any attention
implementation (FlashAttention versions, xFormers, custom kernels);
\item the cap yields \code{excess\_mean}, a \emph{real-time counterfactual diagnostic} --- the patched
softmax is an ``after-the-fact fix'' and cannot tell you ``how this step would have blown up'';
\item the cap keeps $\lambda$ in a safe range while continuing to learn new directions, so training
settles into a stable state (a peaked attention).
\end{enumerate}

\subsection{Figures}
\label{sec:fa:figs}

\begin{figure}[htbp]
\centering
\includegraphics[width=0.85\textwidth]{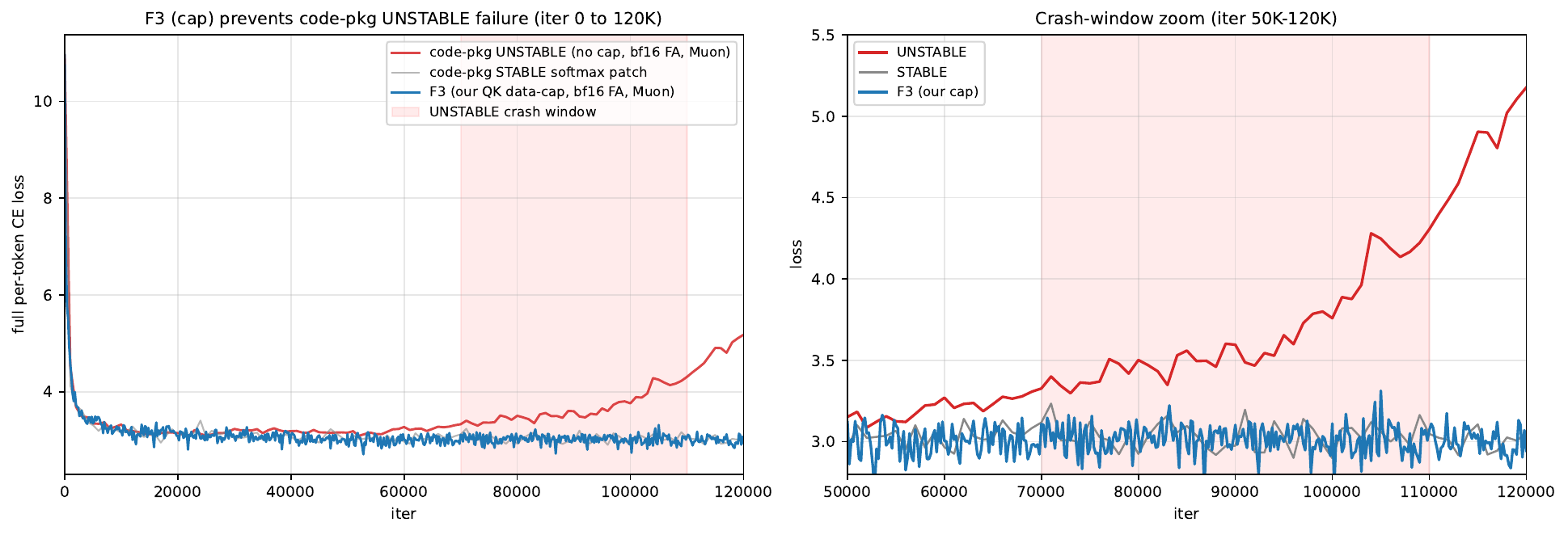}
\caption{\textbf{Loss trajectory: F3 vs uncapped vs patched} (horizontal axis up to 120K, the steps we
actually ran F3 for). Blue $=$ F3 (ours, with the QK cap and no softmax patch); red $=$ uncapped
(no cap, no patch, known to crash); grey $=$ patched (no cap, with the softmax patch). The red shading
$=$ the 70K--110K crash window. F3 (blue) traverses the whole crash window with loss stable at
$3.0$--$3.1$; the uncapped run (red) begins creeping at 80K, crashes at 104K (loss $4.28\to 4.49$), and
reaches 5.18 by 120K. \emph{F3 vs uncapped is a strictly controlled comparison}: same 327K tok/iter,
same Muon, same GPT-2S, same data, same lr schedule --- the only variable is whether the QK covariance
cap is on.}
\label{fig:fa:loss}
\end{figure}

\begin{figure}[htbp]
\centering
\includegraphics[width=0.85\textwidth]{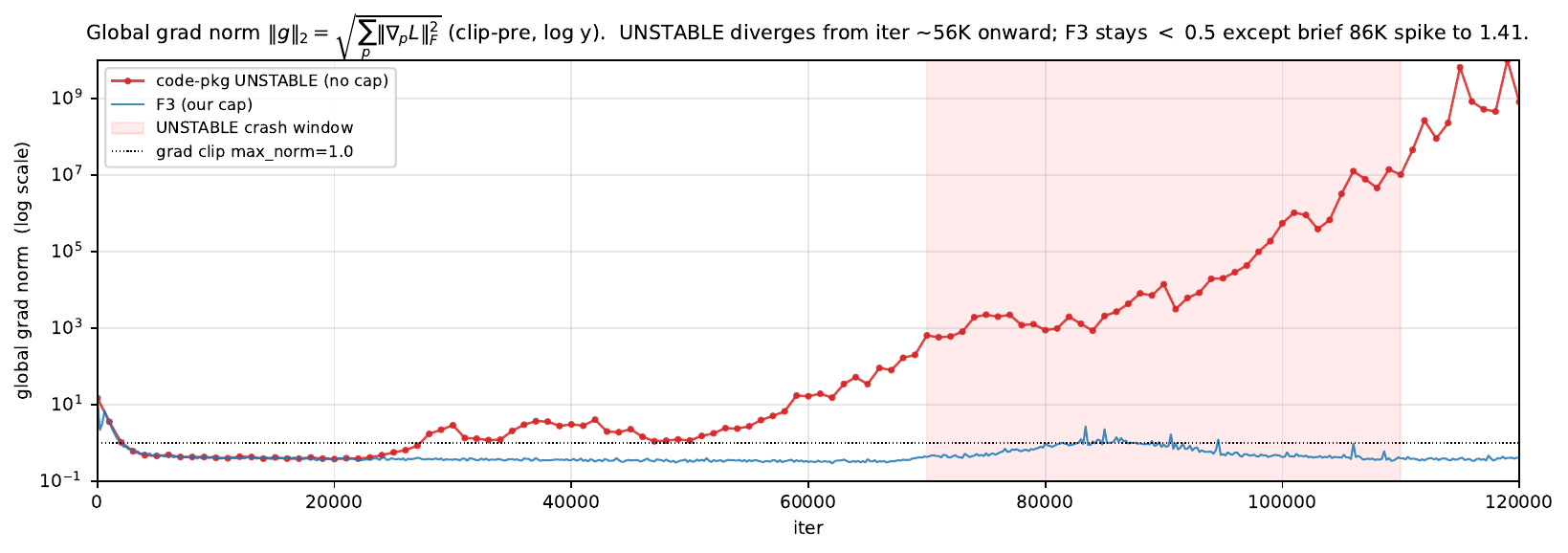}
\caption{\textbf{Global grad-norm trajectory} (log-$y$ axis, up to 120K). The global grad norm
$\|g\|_2 = \sqrt{\sum_p \Fro{\nabla_p L}^2}$ (over all parameters $p$, before clipping), one value per
train step. Blue $=$ F3 (continuous trace); red dots $=$ the uncapped baseline (one summary point per
1000 iters). Dashed black $=$ grad-clip max-norm $= 1.0$. The two overlap before 56K
(grad norm $\sim 0.4$--$0.5$); from $\sim 56$K the uncapped run explodes --- 64K$=$52, 72K$=$609,
80K$=$886, 88K$=$8K, 96K$=$29K, 104K$=6.7\times 10^5$, 120K$=8.1\times 10^8$ --- eight orders of
magnitude. F3 stays $< 0.5$ throughout (except an iter-86K spike to 1.41 that falls back within
seconds, the moment the L1 H2 cap acts). A log $y$ axis is needed to see the full uncapped trajectory.}
\label{fig:fa:grad}
\end{figure}

\begin{figure}[htbp]
\centering
\includegraphics[width=0.85\textwidth]{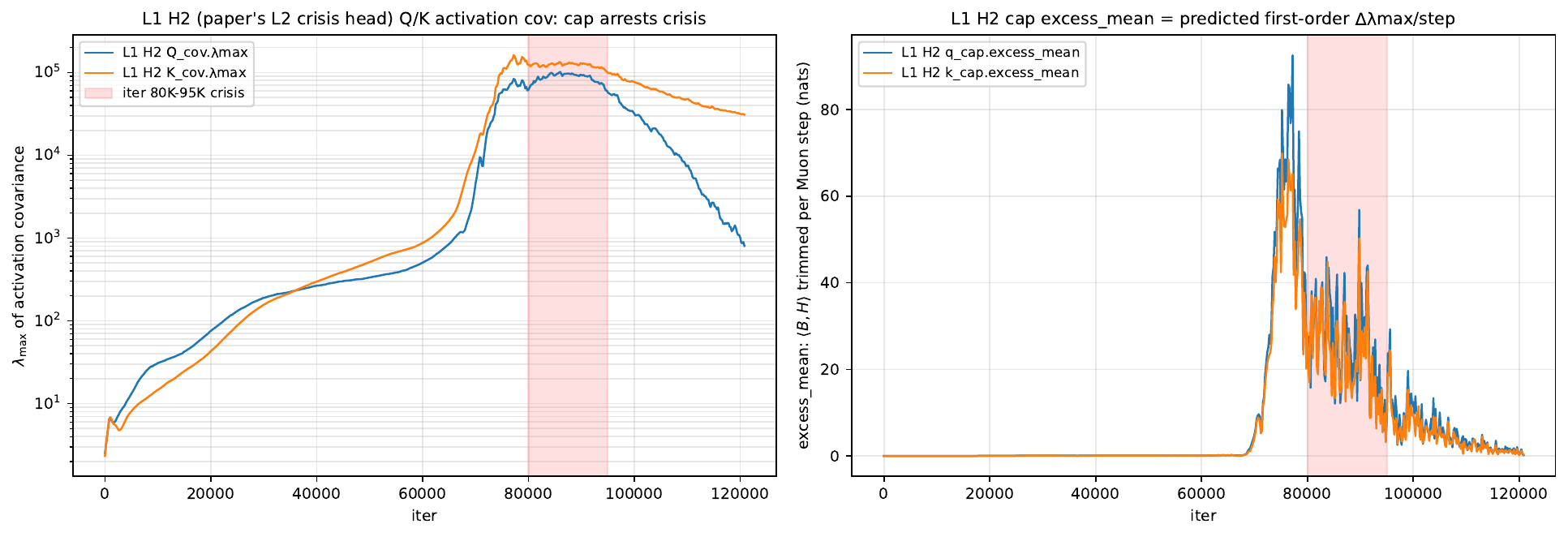}
\caption{\textbf{The 86K crisis at L1 H2} (this figure is \emph{specifically} for L1 H2). Left axis:
$\lambda_{\max}(K_Q^{L1, H2})$ (blue) and $\lambda_{\max}(K_K^{L1, H2})$ (red) over the run; right axis:
the corresponding q\_excess (green) and k\_excess (orange). Red shading $=$ the 80K--95K crisis window.
$\lambda_K$ rises from 873 at 60K to a peak of 133K at 86K ($150\times$), and the cap excess rises in
step from 0.13 to 22/step ($170\times$). The cap intercepts hardest exactly at the $\lambda$ peak.}
\label{fig:fa:L1H2crisis}
\end{figure}

\begin{figure}[htbp]
\centering
\includegraphics[width=0.85\textwidth]{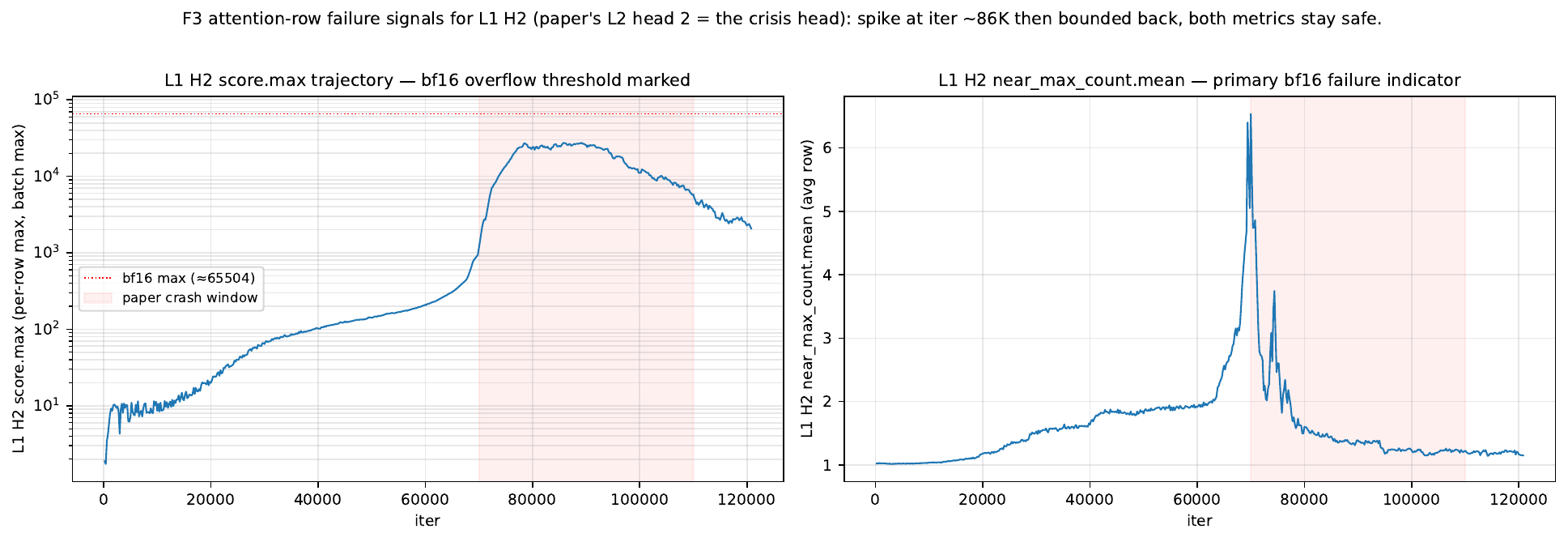}
\caption{\textbf{Two bf16-safety quantities of the L1 H2 attention scores}. Left: \code{score.max}
(the single largest score entry over $(B, T_q, T_k)$ for this head, not a row-max average, log scale);
right: \code{near\_max.mean} (the number of ``near row max'' entries per query token, averaged over
batch and query). The bf16 max, 65504, is marked in red. Both spike at iter 86K then recover:
score.max $209 \to 27{,}264 \to 2{,}272$; near\_max.mean $1.9 \to 6.5$ (70K) $\to 1.4$ (86K, attention
now peaked) $\to 1.2$ (120K). Neither breaks the bf16 mantissa threshold, so the softmax error does not
accumulate.}
\label{fig:fa:scorenear}
\end{figure}

\subsection{Part 4 summary}
\label{sec:fa:sum}

\begin{itemize}\itemsep0pt
\item F3 fully covers the uncapped baseline's crash window (70K--110K) with a stable loss
(Fig.~\ref{fig:fa:loss});
\item L1 H2 undergoes a real near-explosion at iter 86K ($\lambda_K = 133$K, score $= 27$K, near half
the bf16 ceiling), and the cap intervenes in real time with excess $= 22$ nats/step;
\item the cap's counterfactual diagnostic \code{excess\_mean} quantifies, in real time, ``what the next
step would have done'' --- information the patched softmax cannot provide;
\item the rescue is a combination of three mechanisms: (i) the cap trims the first-order growth rate;
(ii) the training dynamics' own attention-peaking lowers near-max; (iii) the cap keeps redirecting $W$
away from the high-$\lambda$ basin.
\end{itemize}

\clearpage

\section{Conclusions across the three case studies}
\label{sec:meta}

\begin{center}
\small
\begin{tabular}{p{0.08\textwidth}p{0.16\textwidth}p{0.18\textwidth}p{0.20\textwidth}p{0.28\textwidth}}
\toprule
Part & $W$ (capped) & $X$ & $K = WK_X W^\top$ & Failure / goal \\
\midrule
2 nano FFN & \code{mlp.c\_proj} $384 \times 1536$ & GELU hidden $\mathbb{R}^{1536}$ & FFN out covariance $\mathbb{R}^{384\times 384}$ & isotropy, PR $\uparrow$ \\
3 MoE router & router $64 \times 384$ & RMSNorm token $\mathbb{R}^{384}$ & sel-score cov $K_{z+b}\in\mathbb{R}^{64\times 64}$ & rank-1 collapse as top-share $\to 1$ \\
4 FA Q/K & per-head $64\times 768$ & layer input $\mathbb{R}^{768}$ & per-head $K_Q^{(\ell,h)}\in\mathbb{R}^{64\times 64}$ & $\lambda_{\max}$ blow-up $\Rightarrow$ bf16 softmax error \\
\bottomrule
\end{tabular}
\end{center}

\textbf{The unified framework (Section~\ref{sec:theory})}:
\begin{enumerate}\itemsep0pt
\item in the scale-invariant regime Muon removes the $1/\Fro{W}$ brake $\Rightarrow$ the
Frobenius/spectral norm drifts more strongly than under SGD;
\item the spectral cap removes the first-order top mode but keeps a non-negative second order
$\Rightarrow$ the weight still learns in new directions;
\item the spectral cap $=$ an $H_\infty$ entropy cap; controlling $H_2$/$H_1$ would need other
projection directions.
\end{enumerate}

\textbf{The unified experimental observations}:
\begin{enumerate}\itemsep0pt
\item $K_X$ has a different dimension and physical meaning in each application, but the cap algorithm
is always ``sample-estimate $K_Y$ + power iteration + first-order projection'', at negligible cost
$O(M\cdot\text{small})$;
\item a data-coupled cap (directly on $K_Y$) is more on-target than a pure $W$-cap (nanoGPT);
\item the cap is decisive at the margins / in crises (MoE L0 rank-1, FA L1 H2 at 86K) and, in the
typical case, leaves the result unchanged while protecting the geometry.
\end{enumerate}

\begin{caveatbox}
\textbf{Limitations and open questions.} (i) The whole theoretical story rests on
Assumption~\ref{ass:scale} (exact scale invariance); the quantitative predictions --- the $1/\Fro{W}$
brake, the $t^{1/4}$ vs $t^{1/2}$ norm growth, and $\langle W, G\rangle_F \approx 0$ --- should be
tested much more broadly than the single checkpoint of Section~\ref{sec:finding1:cpcg} before being
relied upon. (ii) The experiments are at small scale (nanoGPT / GPT-2 small, tens of thousands of
steps) and single-seed in places; the effects at larger scale are untested. (iii) The FlashAttention
result compares against one uncapped log rather than a matched ensemble. We regard these as a coherent
first set of observations rather than a settled account, and we welcome comments, corrections, and
counter-examples.
\end{caveatbox}

\clearpage
\appendix

\section{Figure index}
\label{app:figures}

\begin{center}
\small
\begin{longtable}{lp{0.6\textwidth}}
\toprule
Path & Content \\
\midrule
\code{nano\_D\_val\_loss.pdf} & nanoGPT D0/D1/D1d val loss (Fig.~\ref{fig:nano:loss}) \\
\code{nano\_D\_isotropy\_pr.pdf} & nanoGPT FFN $K_Y$ participation rank (Fig.~\ref{fig:nano:pr}) \\
\code{g3\_val\_loss\_3way.pdf} & MoE G0/G1/G3 val loss (Fig.~\ref{fig:moe:val}) \\
\code{g3\_per\_layer\_cv\_3way\_seed42.pdf} & MoE seed=42 per-layer load CV (Fig.~\ref{fig:moe:cv}) \\
\code{g3\_per\_expert\_load\_3way\_seed42.pdf} & MoE seed=42 per-expert load histogram (Fig.~\ref{fig:moe:bar}) \\
\code{fa\_F3\_loss\_vs\_paper.pdf} & F3 vs baseline loss trajectory (Fig.~\ref{fig:fa:loss}) \\
\code{fa\_F3\_grad\_norm.pdf} & F3 grad-norm trajectory (Fig.~\ref{fig:fa:grad}) \\
\code{fa\_F3\_L2\_crisis.pdf} & L1 H2 86K $\lambda_{\max}$ + excess crisis (Fig.~\ref{fig:fa:L1H2crisis}) \\
\code{fa\_F3\_score\_nearmax.pdf} & L1 H2 score.max + near\_max (Fig.~\ref{fig:fa:scorenear}) \\
\bottomrule
\end{longtable}
\end{center}

\end{document}